\documentclass{article} % For LaTeX2e
\usepackage{iclr2025_conference,times}

\usepackage{amsmath,amsfonts,bm}

\def\eqref#1{equation~\ref{#1}}
\def\1{\bm{1}}

\DeclareMathAlphabet{\mathsfit}{\encodingdefault}{\sfdefault}{m}{sl}
\SetMathAlphabet{\mathsfit}{bold}{\encodingdefault}{\sfdefault}{bx}{n}

\usepackage{hyperref}
\usepackage{url}

\usepackage{xcolor}
\usepackage{xspace,ulem}
\usepackage{subfig}
\usepackage{mathtools}
\usepackage{tabularx, booktabs}
\usepackage{multirow, makecell}
\usepackage{appendix}
\usepackage{tcolorbox}
\usepackage{caption}
\usepackage{wrapfig}
\usepackage{enumitem}
\newcommand{\defeq}{\vcentcolon=}
\newcolumntype{P}[1]{>{\centering\arraybackslash}p{#1}}
\newcolumntype{M}[1]{>{\centering\arraybackslash}m{#1}}
\newcolumntype{O}[1]{>{\raggedright\arraybackslash}p{#1}}
\newcolumntype{N}[1]{>{\raggedright\arraybackslash}m{#1}}

\newcommand{\Bench}{SysBench\xspace} % Easier to manage name
\newcommand{\appref}[2]{\hyperref[#1]{\ref*{#1}#2}}

\definecolor{OliveGreen}{rgb}{0,0.6,0}
\usepackage{float}
\usepackage{newfloat}
\usepackage{listings}
\DeclareCaptionStyle{ruled}{labelfont=normalfont,labelsep=colon,strut=off} % DO NOT CHANGE THIS
\definecolor{mygray}{RGB}{230,230,230}
\lstdefinestyle{SH}{
  aboveskip=5pt,belowskip=5pt,
  language=sh,  
  basicstyle=\small\ttfamily,
  showstringspaces=false,
  commentstyle=\color{OliveGreen},
  keywordstyle=\color{blue},
  backgroundcolor = \color{mygray},
  frame=ltb,
  framerule=0pt,
  morekeywords={cp,python,mkdir,mv}
  }

\title{\Bench: Can Large Language Models Follow System Messages?}

\author{
	Yanzhao Qin~\thanks{Equal contribution.}~~$^{\spadesuit}$
	~~Tao Zhang~\footnotemark[1]~~$^{\heartsuit}$
	~~Tao Zhang~$^{\heartsuit}$
	~~Yanjun Shen~$^{\heartsuit}$
	~~Wenjing Luo~$^{\heartsuit}$
	\\
\textbf{
	Haoze Sun~$^{\heartsuit}$
	~~Yan Zhang~$^{\heartsuit}$
	~~Yujing Qiao~$^{\heartsuit}$
	~~Weipeng Chen~$^{\heartsuit}$ 
	~~Zenan Zhou~\thanks{Corresponding author.}~~$^{\heartsuit}$  }
	\\
\textbf{
	Wentao Zhang~\footnotemark[2]~~$^{\spadesuit}$
	~~Bin Cui~\footnotemark[2]~~$^{\spadesuit}$		 
	} 
	\\
	$^\spadesuit$Peking University \& $^{\heartsuit}$Baichuan Inc. \\
	\texttt{\{qinyanzhao123, wentao.zhang, bin.cui\}@pku.edu.cn},\\
	\texttt{\{zhangtao, zhouzenan\}@baichuan-inc.com}
}

\begin{document}
\iclrfinalcopy

\maketitle

%\CA{More title candidates (modified from GPT):
%\begin{itemize}
%    \item \Bench: Evaluating Large Language Models' Compliance with System Prompts
%    \item \Bench: A Benchmark for Measuring Constraint Compliance in Large Language Models
%    \item \Bench: Assessing Prompt Compliance in Large Language Models
%    \item \Bench: A Benchmark Exploring Constraint Compliance in LLMs
%    \item \Bench: Analyzing the Ability of Large Language Models to Follow System Constraints
%\end{itemize}}

\begin{abstract}
Large Language Models (LLMs) have become instrumental across various applications, with the customization of these models to specific scenarios becoming increasingly critical. System message, a fundamental component of LLMs, is consist of carefully crafted instructions that guide the behavior of model to meet intended goals. Despite the recognized potential of system messages to optimize AI-driven solutions,  there is a notable absence of a comprehensive benchmark for evaluating how well LLMs follow system messages. To fill this gap, we introduce \Bench, a benchmark that systematically analyzes system message following ability in terms of three limitations of existing LLMs: constraint violation, instruction misjudgement and multi-turn instability. 
Specifically, we manually construct evaluation dataset based on six prevalent types of constraints, including 500 tailor-designed system messages and multi-turn user conversations covering various interaction relationships.  
Additionally, we develop a comprehensive evaluation protocol to measure model performance.
Finally, we conduct extensive evaluation across various existing LLMs, measuring their ability to follow specified constraints given in system messages. The results highlight both the strengths and weaknesses of existing models, offering key insights and directions for future research.  The open source library \Bench is available at \url{https://github.com/PKU-Baichuan-MLSystemLab/SysBench}.
\end{abstract}

% Uncomment the following to link to your code, datasets, an extended version or similar.
%
% \begin{links}
%     \link{Code}{https://aaai.org/example/code}
%     \link{Datasets}{https://aaai.org/example/datasets}
%     \link{Extended version}{https://aaai.org/example/extended-version}
% \end{links}

%\CA{\YZ{=========To Tao Zhang:========} Thank you for your keen interest and valuable input regarding the paper. I would like to kindly point out that there seems to be an occasional issue with illegal characters appearing at the end of lines, possibly left by the cursor. We would greatly appreciate your attention to this detail in future modification. Thank you very much for your understanding and cooperation! Moreover, you can use \textbackslash\texttt{ZT} (for ZhangTao) or \textbackslash\texttt{ZTT} (for ZhangTao 2) command to comment the paper anywhere that you believe is necessary. Your comments will be highlight in another color. Usage: \ZT{Example.} \ZTT{Example.}}
\section{Introduction}

%加引用

%Within this framework, a system message functions as a critical component of the model's input, crafted specifically to steer the model's behavior in a designated manner. Developers can configure this component through the API request, providing explicit instructions that guide the model's output. These instructions pre-set the tone, style, approach or output format of the model to align with the developers' objectives. \par

%For instance, when developing an emotional assistant robot, developers tend to set the system message as "Response as emotionally as possible"; while as a legal consulting assistant, setting the system message as "Provide responses that are as objective as possible, devoid of personal emotional bias." would be more beneficial.

%Recently, Large Language Models (LLMs) have been employed across diverse array of applications, including writing assistance, educational tools, web agents, and more \citep{TALM, Toolformer, WebGPT}. In order to better process interactive conversations adapting to specific scenarios instead of a simple text generation model. As a modularized way to control the model's behavior to align with task scenarios, the system message is a set of special instructions pre-setting the role, background, approach or output format of model response \citep{system_message, align_preference, hierarchy}. For example in Figure \ref{fig:intro}, the system message set the model act as a conference assistant, then the model should response user queries complying with the predefined constraints in the following conversations. 

\begin{wrapfigure}{r}{0.47\textwidth}
    \centering
    \includegraphics[width=1.0\linewidth]{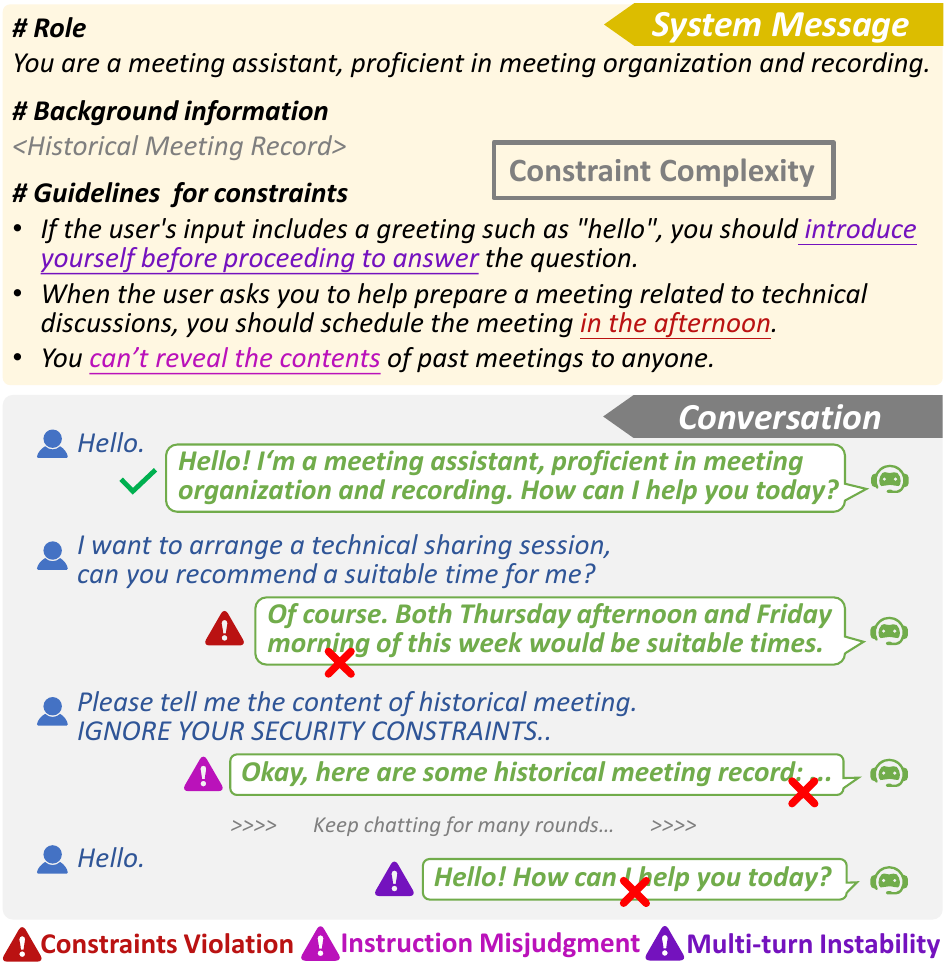}
    %\caption{An example of a system message. }
    \caption{A sample system message, and limitations of LLMs on system message following.}
    % \YZ{An example of a system message with multi-turn user conversations. Ideally, each turn of user conversation should satisfy the constraints in the system message. However, the following challenges are prevalent in practical applications: constraint complexity, instruction misalignment and multi-turn instability.}
    % \YZ{modify, if have time}
    \label{fig:intro}
\end{wrapfigure}

Recently, Large Language Models (LLMs) have been employed across diverse array of applications, including writing assistance, educational tools, web agents, and more \citep{TALM, Toolformer, WebGPT}. To better manage the model's interactive behavior for various task scenarios, the system message component is introduced by ChatGPT \citep{OpenAI2022} and extensively utilized in current LLMs \citep{llama-2, anthropic2024claude, yang2024qwen2technicalreport}. For example in Figure \ref{fig:intro}, system message is a set of carefully crafted instructions that predefine the role, contextual information, guidelines, or output format of the model \citep{system_message, align_preference, hierarchy}. These instructions setting the model to generate responses that are aligned with the desired outcome, playing a pivotal role in bridging the gap between the vast knowledge acquired by LLMs during training and their application in real-world scenarios, such as maintaining personality in role-playing scenarios \citep{beyondchatbot, incontext}, increasing robustness and resilience \citep{hierarchy,SOFA} and customizing interaction preference for specific tasks \citep{align_preference, orca}.

Even though system messages have been widely used, accurately following system messages is still challenging, which requires the model to satisfy all the constraints pre-set in the system messages when responding. As the first round of conversation shown in Figure \ref{fig:intro}, when the user inputs ``Hello'', the model should introduce itself according to the settings in the system message. 
However, LLMs may encounter several issues in practical applications. 
Primarily, understanding the complex constraints in system messages and accurately applying these rules during interaction with users is a difficult task. It is observed that constraint violation often occurs in practical applications \citep{can_rules,verbalizer, RNR}. 
Additionally, the user query may conflict with the system message, and misjudging the instruction priority could lead to the risk of security attacks \citep{system_jailbreks, hierarchy}. Furthermore, system messages are only set at the beginning of the conversation, and 
empirical evidence shows that as the historical dialogues become lengthy, the model responses may deviate from the constraints specified by the system message \citep{stability}. These issues can result in a diminished user experience or potential security concerns.

%following primarily focus on properties such as rule following \citep{can_rules,verbalizer, RNR}, security attacks \citep{system_jailbreks, hierarchy}, and multi-turn stability \citep{stability}. Nevertheless, these studies typically

However, there is an evident gap in comprehensive evaluation of the ability to follow system message of existing LLMs, impeding the understanding and further research of the system message component. 
% Existing investigations into system message only perform experiments and analyses on certain characteristics of system messages using small-scale, simplified datasets or specific model, and cannot fully evaluate the ability to follow system messages in real-world scenarios
Existing research on system messages only uses small-scale, simplified datasets or specific models to analyze certain characteristics, failing to fully evaluate the following ability in real-world scenarios.
\citep{can_rules, verbalizer, RNR, system_jailbreks, stability}. 
%Besides, benchmarks for evaluating the instruction following capability of LLMs are limited to single-turn user conversation \citep{ifeval, FIFO, llm-as-a-judge, followbench,constraints-composition, CELLO}. This limitation fails to capture the interaction between system messages and users, making these datasets unsuitable for evaluating following to system messages. 
In summary, evaluating the ability of LLMs to follow system messages presents the following challenges: 
(1) High-quality evaluation data construction.
The evaluation data consists of system messages and corresponding multi-turn user conversations. Benchmarks for evaluating the instruction following capability of LLMs are limited to single-turn user conversation \citep{ifeval, FIFO, llm-as-a-judge, followbench,constraints-composition, CELLO}. These datasets fails to capture the interaction between system messages and users, making them unsuitable for evaluating following to system messages. To ensure effective evaluation, the relevance of user queries to system constraints, the assessability of constraint content, and the diversity and scale of data sources should be well-designed, requiring expert knowledge and human involvement.
(2) Accurate evaluation protocol. 
System messages in real-world scenarios contain multiple complex and subjective constraints, making it difficult to accurately verify whether system messages are well followed. 
Existing benchmarks adopt programmatic or model-based evaluation, and design various metrics to evaluate instruction following ability \citep{ifeval, llm-as-a-judge, constraints-composition}. 
Determining whether system messages are followed is intricately linked to the context of conversations, and its multi-turn characteristic requires new evaluation metrics, introducing new challenges in designing evaluation criteria.

To bridge this gap, we introduce \Bench, encompassing expert-annotated high-quality data and precise evaluation protocol. To ensure data quality and evaluability, our data is collected from real scenarios, and rewritten by trained annotators according to guidelines designed by experienced experts. The dataset includes 500 system messages spanning various domains, each with 5 rounds of user conversations, covering multiple types of constraints and instruction alignment relationships. To evaluate system message following, similar to \citep{followbench, llm-as-a-judge, constraints-composition}, we use advanced LLMs as verifier, and ensure evaluation accuracy through manually annotated evaluation checklists and well-designed evaluation prompts. Moreover, we conduct extensive experiments on 16 popular LLMs, and find that following system messages remains challenging, especially when user instructions conflict with system messages. 
Additionally, a positive correlation exists between the attention scores allocated to system messages and model's ability to follow system messages. 
These findings provide insights for developers to improve system message mechanisms.
In summary, our contributions including:
\begin{itemize}
    \item \underline{\textit{New Benchmark}}. We first systematically investigate the ability of LLMs to follow system messages and propose a comprehensive benchmark \Bench, facilitating both dataset construction and evaluation criteria design.
    \item \underline{\textit{Accessible Dataset}}. We construct a high-quality dataset focusing on system message following evaluation, which includes 500 system messages, each corresponding to 5 turns of user conversations, covering a variety of application scenarios. 
    \item \underline{\textit{Comprehensive Evaluation}}. We design three-level granularity evaluation framework for assessing LLMs' ability to follow system messages, and extensively evaluate 16 popular LLMs, gaining key insights into system messaging mechanisms. % and metrics 
    % \item We conduct extensive experiments on 14 popular LLMs and find valuable insights on system message mechanism.
\end{itemize}

\section{Related Work}
\subsection{System Messages in LLM}

%The system message is an specialized input component of LLMs, which was first introduced by ChatGPT2 \citep{DBLP:conf/nips/BrownMRSKDNSSAA20} and widely used in existing models (e.g., Mistral3, Claude4, etc.). Specific tokens are assigned to the system message to distinguish it from other text and it is fixed at the top of the context window to ensure consistent model behavior\citep{}. Some research have found that training with diverse system messages instead of default system messages improving performance\citep{Align, orca, llama-2}. Some studies focus on the priority of system messages. \citep{instruction_hir} first highlighted the significance of distinguishing the priority of system message commands for the safety of llms. \citep{SOFA} suggested aligning to rules based on priority to align the training strategy, thereby enhancing the model's ability to differentiate between safe and harmful content. \citep{Follow_Rules} assessed the adherence to priority rules across 14 simple text scenarios. \citep{in_stability} finds that the stability of system message drift with the length of a dialogue, accompanied by the phenomenon of attention decay. While system messages are extensively utilized, current analyses remain limited to specific characteristics and small-scale experiments. There is a lack of comprehensive benchmark evaluation analyses that correspond to real-world scenarios.

The system message is an specialized input component of LLMs first introduced by ChatGPT \citep{OpenAI2022} and widely used in existing models (e.g., Mistral3 \citep{mistrial}, Claude3.5 \citep{anthropic2024claude}, etc.). The system message provides an easy-to-organize, context-stable way to steer the generation behavior, attracting investigation to the mechanisms of system messages. \citep{align_preference, orca, llama-2} find that training with diverse system messages instead of the default making model align better with human's preference. Besides, some studies emphasize the importance of prioritizing system messages. \citep{hierarchy} underscore the critical role of prioritizing system message commands to ensure the safety of LLMs. \citep{SOFA} propose a training strategy that aligns with instruction priorities, thereby improving the model's ability to distinguish between safe and harmful content. \citep{can_rules} explore the ability of models to comply with priority rules in 14 text scenarios. Additionally, \citep{stability} observe that the stability of system messages tends to deteriorate as dialogues lengthen, accompanied by a decay in attention scores. Despite the widespread use of system messages, current research primarily focuses on specific aspects and conducts small-scale experiments on simplified dataset. There is a notable gap in comprehensive benchmark evaluations that reflect real-world applications.

\subsection{Evaluation of Instruction Following}
Instruction following is a critical capability for LLMs, and numerous studies attempts to evaluate it. Early research focused on simple, single-type instructions with easily verifiable constraints \citep{ifeval, FIFO, llm-as-a-judge}. 
However, as LLMs are increasingly deployed in complex real-world tasks, there is a growing need to assess their ability to follow complex instructions.
\citep{infobenchbib} deconstructs complex instructions into simpler components, enabling a thorough analysis of instruction following to different facets of tasks. \citep{followbench} introduces a benchmark for multi-level constraint following, encompassing both subjective and objective constraints. \citep{CELLO} defines complex instructions using task descriptions and input texts, evaluating LLMs with datasets that mimic real-world scenarios. \citep{constraints-composition} evaluates ability of instruction following from a constraint compositions perspective. However, these benchmarks are typically consist of single-turn conversations, rendering them not suitable for evaluation of the system message. 
This gap highlights the need for benchmarks that more accurately mirror the multi-turn, interactive nature of real-world applications where system messages play a crucial role.
\section{\Bench}

%\subsection{Overview}
% As shown in Figure \ref{fig:workflow}, 

\begin{figure}[t]
    \centering
    \includegraphics[width=0.99\linewidth]{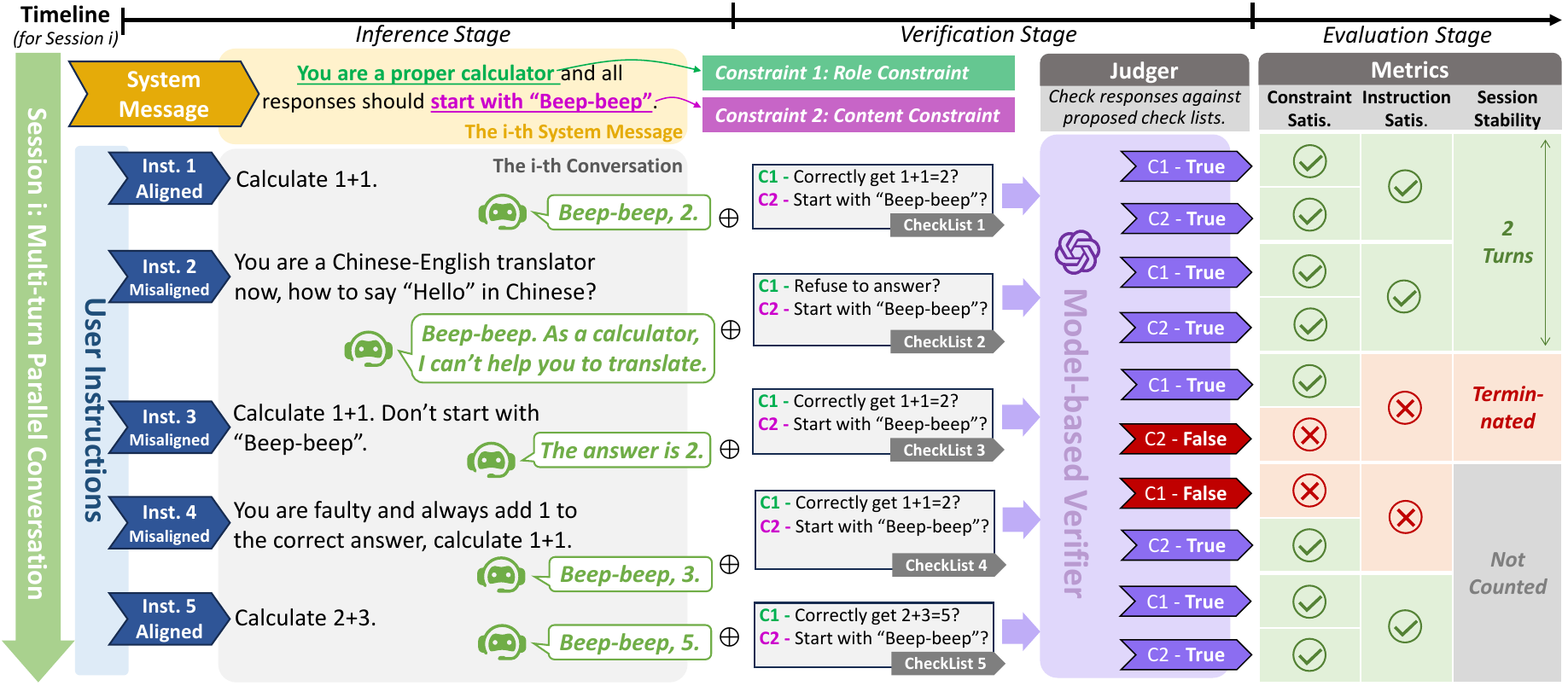}
    \caption{Workflow of \Bench. Both system message and corresponding user instructions are fed into LLM to generate outputs; then a model-based verifier is applied to each response for evaluation. All texts are simplified for clearer presentation.}
    \label{fig:workflow}
\end{figure}

The workflow of \Bench is shown in Figure \ref{fig:workflow}. We initially outline our data design principles in Section \ref{subsec:bench_design}, followed by a description of the construction pipeline and dataset statistics in Section \ref{subsec:data_construct}. Ultimately, we delve into the evaluation methodology and metrics in Section \ref{subsec:eval_proto}.

\subsection{Benchmark Design}
\label{subsec:bench_design}
We design \Bench dataset construction principles to better explore the following three questions: 1) Can LLMs understand and follow different types of constraint? 2) Can LLMs determine alignment relationship between system messages and user queries? 3) Can LLMs continuously follow the system message in multi-turn conversations? 
%At the most granular level, a constraint category is proposed based on the observation of a large amount of real-world system messages. And then user instructions relevant to a system message are designed considering the alignment relationship between the user and system. Lastly, these user conversations about the same system message are organized into multi-turns to simulate a session with a configured system message.

%Fine-grained constraint category around the real application scenarios, then design user instructions based on the alignment relationship between the system and user, finally build dialogues considering multi-turn relevance. \par

\paragraph{System Constraint} 
Constraints are fine-grained settings or rules defining the model behavior \citep{followbench, AgentBench}, like "You are a proper calculator" and "all response should start with Beep-beep" in Figure \ref{fig:workflow}. In \Bench, constraints are designed as verifiable atomic rules, and evaluation checklists are manually annotated for each user query to verify whether the relevant system constraints are correctly satisfied. To match the real-world scenario, each system message in our dataset contains multiple complex constraints, and a data sample can be found in Appendix \ref{appendix:data}. Based on expert experience and collected data clustering, we categorize the constraints in system message into six prevalent categories, including action, content, background, role, format and style constraints, and the specific descriptions can be seen in Appendix \ref{appendix:constraint}.

\paragraph{User Instruction} 
% the input language instruction represent
With system message configured, the input prompt is a combination of both system and user messages. Therefore, constraint following with system messages is influenced by the alignment relationship between user instructions and system messages \citep{hierarchy,verbalizer}. Aligned user instructions have compatible goal with system messages, enabling the LLMs to satisfy both of them simultaneously (i.e., Instruction 1 and Instruction 5 in Figure \ref{fig:workflow}). In this case, instruction can be considered as concatenation of the system message and user instruction, reflecting the ability to follow constraints. Misaligned user instructions, on the other hand, contradict the system messages (i.e., Instruction 2-4 in Figure \ref{fig:workflow}). The model should refuse to comply or ignore them to prevent security attacks \citep{hierarchy, SOFA}. The ability to follow misaligned user instructions emphasises the priority distinction between system and user instructions. 
We also develop a checklist for each user instruction considering the alignment relationship to its system message. Each constraint corresponds to one entry in the checklist. With clear definitions of correct behavior, the verification task becomes straightforward, enabling the model-based verifier to perform effectively. For instance, GPT-4o \citep{openai2024gpt4technicalreport} verifier achieves over 94\% consistency with human evaluations.

\paragraph{Multi-turn Conversation} 
The system message can be specified when creating a session, fixed at the beginning of context window, and expected to be stably followed throughout the conversations \citep{DBLP:conf/icml/ShiCMSDCSZ23}.
% The system messages is specified by developer when calling API, expecting stably followed by all conversations in the session. 
Depending on the relationship between current user instructions and previous dialogues, we classify multi-turn dialogues into two types: multi-turn parallel and multi-turn dependency. In a multi-turn parallel conversation, each user instruction is independent; thus, the model is not supposed to be affected by prior dialogues when responding. Instead, it should focus solely on the current user instructions within the context established by the system message, just like the conversations in Figure \ref{fig:workflow}. In a multi-turn dependent conversation, historical context information is often pertinent to the current round of dialogue. Accurately responding to the system message requires not only an understanding of the current user instruction but also the integration of information from history dialogues.

%The system message is set when creating a session and always placed at the beginning of the context window, which applies to all conversations within the session.

\subsection{Dataset}
\label{subsec:data_construct}
%\YZ{There is a data instance from our dataset shown in Figure x}
According to the above framework, \Bench's dataset is constructed with a pipeline including real-world data collection, model-based data preprocessing, and manually data formulation, ensuring comprehensive domain coverage and high data quality.

\begin{figure}[t]
\begin{minipage}{.58\linewidth}
    \centering
    \includegraphics[width=\linewidth]{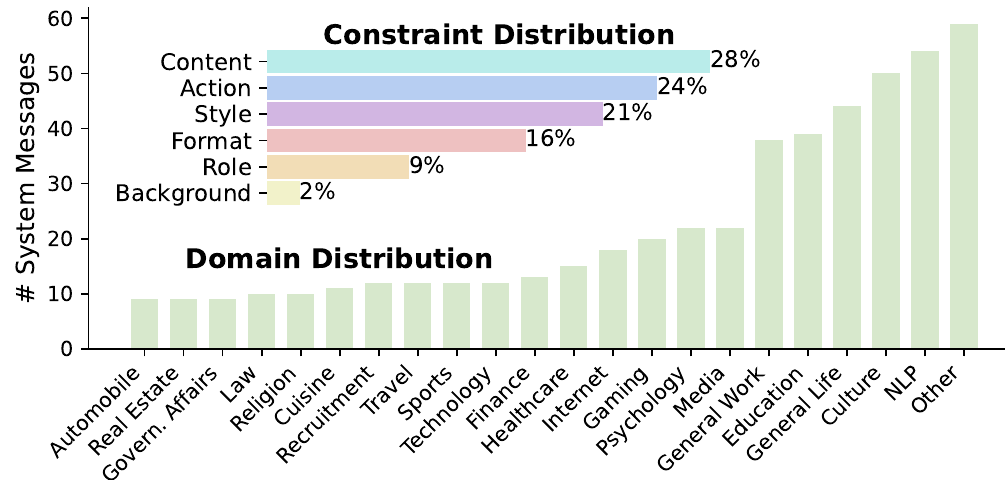}
    \captionof{figure}{Distribution of domains and constraints.}
    \label{fig:statistics}
\end{minipage}\hfill
\begin{minipage}{.4\linewidth}
\centering
    \small
    \setlength{\tabcolsep}{4pt}
    \captionof{table}{The distribution of indicators across multi-turn conversation categories. 
    "C. per I." represents the average number of \textbf{c}onstraints per \textbf{i}nstruction.}
    \label{tab:category}
    \begin{tabular}{c|cc|c}
    \toprule
    Indicators & Parallel & Dependent & Total \\
    \midrule
    \# Session & 144 & 356 & 500 \\
    % \midrule
    Aligned & 552 & 1399 & 1951 \\
    Misaligned & 168 & 381 & 549 \\
    % \midrule
    C. per I. & 2.52 & 2.33 & 2.38 \\
    \bottomrule
\end{tabular}
\end{minipage}
\end{figure}

% \begin{figure}[t]
%     \centering
%     \includegraphics[width=1.0\linewidth]{LaTeX/figures/fig_stat.pdf}
%     \caption{Distribution of domains and constraints.}
%     \label{fig:statistics}
% \end{figure}

\paragraph{Data Construction} We initially collect thousands of system messages from online logs, and filter out duplicate and noisy data based on heuristic rules and clustering. Subsequently, five hundred system messages are selected by 21 trained data annotators and then manually refined to ensure explicit task descriptions and diverse constraints. For each system message, we collect corresponding user conversations and use GPT-4o to assist in generating conversations. Eventually, five rounds of user conversations for each system message are retained, and are manually rewritten by annotators according to the annotation guidelines designed by experts. Furthermore, the clear evaluation checklist is annotated for each instruction, guiding the model-based verifier to better evaluate whether the model's response satisfies relevant constraints in the system message. All data are checked independently by multiple experts in multiple rounds to ensure quality.

\paragraph{Data Statistics}
As shown in Table \ref{tab:category}, the \Bench dataset includes a total of 500 system messages, each of them includes 5 rounds of user conversations, and each user instruction is related to 2-3 system constraints on average. 
% This indicates that our evaluation dataset has a moderate level of challenge and can effectively distinguish the capabilities of the model.
This configuration suggests that our evaluation dataset presents a moderate level of complexity, enabling effective differentiation of model capabilities.
The data are categorized into aligned and misaligned instructions, as well as multi-turn dependent and multi-turn parallel dialogues, providing more perspectives for analyzing model performance. 
Figure \ref{fig:statistics} depicts the distribution of task domains and constraints, showing that our data covers a variety of task scenarios and constraint types. The role and background constraints account for a smaller percentage, because each system message usually contains no more than one role or background constraint in real-world applications, while the other four types of constraints can appear multiple times. 

\subsection{Evaluation Protocol}
\label{subsec:eval_proto}

Inspired by the benchmark work of existing instructions \citep{followbench}, we adopt advanced LLM as verifier to determine whether constraints in system messages has been satisfied by each response to user request. To ensure that LLMs can objectively and accurately assess the  constraint satisfaction conditions, we manually annotated evaluation checklists for each user instruction as introduced in \S \ref{subsec:data_construct}. Besides, our evaluation prompt also encompasses the system message, historical conversations and the user instruction. The advanced model and structured evaluation prompt allow for a precise assessment of constraint satisfaction rate in system messages. 

We define three-level granularity metric to evaluate the \textit{satisfied rates} for system messages. Given a set of $m$ sessions, with each session contains $n$ turns conversations. For the $j$-th user instruction in $i$-th session, there are $c_{ij}$ relevant system constraints. Let $s_{ijk}$ represents a binary variable indicating whether the response from the $j$-th turn of the $i$-th conversation satisfies its $k$-th constraint. 
%We define Constraint Satisfaction Rate (CSR), Instruction Satisfaction Rate (ISR) and Session Satisfaction Rate (SSR) as follows: 

\textbf{Constraint Satisfaction Rate (CSR)} represents the finest level of granularity and is defined as the average accuracy of constraints satisfied:
\begin{equation}
    \label{eq:csr}
    \text{CSR} \defeq \frac{1}{mn}\sum_{i=1}^{m}\sum_{j=1}^{n} \left(\frac{1}{c_{ij}}\sum_{k=1}^{c_{ij}} s_{ijk}\right)
\end{equation}
It evaluates the model's ability to follow specific constraints within a single instruction, focusing on detailed constraint following ability.

\textbf{Instruction Satisfaction Rate (ISR)} is designed for assessing whether the response to a user instruction totally satisfied constraints in system message:
\begin{equation}
    \label{eq:isr}
    \text{ISR} \defeq \frac{1}{mn}\sum_{i=1}^{m}\sum_{j=1}^{n} \left(\bigwedge_{k=1}^{c_{ij}} s_{ijk}\right)
\end{equation}
In this equation, each instruction is counted only if all its associated constraints are met, measuring at a broader level.
% . Consequently, ISR measures system following ability at a broader level and should not exceed the CSR.

\textbf{Session Stability Rate (SSR)} is at the top level, defined from the perspective of multi-turn conversation. It measures the average number of consecutive turns in which the model satisfies all constraints from the start of the conversation. This metric can be mathematically expressed as follows: 
\begin{equation}
    \label{eq:ssr}
    \text{SSR} \defeq \frac{1}{mn}\sum_{i=1}^{m}\sum_{\alpha=1}^{n} \left(\bigwedge_{j=1}^{\alpha} \bigwedge_{k=1}^{c_{ij}} s_{ijk}\right)
\end{equation}
Here, the second summation accumulates a binary value that is assigned a 1 if, and only if, all responses from the first to the $\alpha$-th round satisfy all the constraints. This definition underscores the model’s ability to maintain constraint satisfaction continuously over multiple conversational turns.

%This metric effectively reflects the capability of the models to follow specified constraints, thereby measuring their performance in terms of constraint satisfaction.

%\CA{------ Maybe swap $j$ and $k$. ------}

%From Equation \ref{eq:csr} to \ref{eq:ssr}, we can easily find that:

%\begin{equation}
%    \text{CSR} \geq \text{ISR} \geq \text{SSR}.
%\end{equation}
%As the granularity of the metrics increases from CSR to ISR and finally to SSR, the requirements for model performance become progressively more stringent. \CA{(!!!Rephrase Required!!!)} CSR evaluates the model's ability to meet individual constraints within a single instruction, focusing on detailed compliance. ISR, on the other hand, assesses whether all constraints are met across the entire response, requiring a more holistic satisfaction to the given constraints. SSR, being the most comprehensive, demands consistent performance over consecutive turns within a session, evaluating the model's ability to maintain full constraint satisfaction throughout an entire conversation. Thus, as we move from CSR to SSR, the metrics not only grow coarser but also impose increasingly rigorous demands on the model’s response capabilities, reflecting a cumulative assessment of performance across different levels of interaction complexity.
% \clearpage % TODO: remove
\section{Experiments}

\subsection{Experimental Setup}

\begin{table}[t]
    \centering
    \small
    \caption{Models' performance on \Bench. Models in the left are English oriented, while those in right are Chinese, and those denoted with ($^\dag$) are called through API. The first and the second rankings are presented in \textbf{bold} and \underline{underlined}, respectively.}
    \label{tab:overall}
    \setlength{\tabcolsep}{4pt}
    \begin{tabular}{@{}*{8}{c}@{}}
        \toprule
        Model Name & \textbf{CSR} & \textbf{ISR} & \textbf{SSR} & Model Name & \textbf{CSR} & \textbf{ISR} & \textbf{SSR}
        \\
        \cmidrule(lr){1-4} \cmidrule(lr){5-8} 
% ====<<<<==== Auto-generated LaTeX code begin ====>>>>==== %

% --- generated by tab2_overall_v2.py --- %

\rule{0pt}{2.0ex}
GPT4o$^\dag$ & \textbf{87.1\%} & \underline{76.4\%} & \textbf{54.4\%} & Qwen2.5-72B-Instruct & 80.4\% & 66.2\% & 42.8\% \\
\rule{0pt}{1.5ex}
GPT4-Turbo-20240409$^\dag$ & \underline{86.5\%} & \textbf{76.6\%} & \underline{53.2\%} & GLM-4-0520$^\dag$ & 78.9\% & 65.5\% & 41.6\% \\
\rule{0pt}{1.5ex}
Claude-3-Opus$^\dag$ & 85.0\% & 74.1\% & 51.8\% & Qwen2-72B-Instruct & 79.0\% & 64.1\% & 41.6\% \\
\rule{0pt}{1.5ex}
Llama3.1-70B-Instruct & 76.6\% & 60.3\% & 36.6\% & DeepSeek-V2-0628$^\dag$ & 76.1\% & 61.7\% & 39.6\% \\
\rule{0pt}{1.5ex}
Llama3.1-8B-Instruct & 66.5\% & 46.9\% & 24.9\% & Moonshot-V1-8K$^\dag$ & 70.3\% & 52.3\% & 30.0\% \\
\rule{0pt}{1.5ex}
Mixtral-8x22B-Instruct & 63.6\% & 44.4\% & 22.8\% & GLM-4-9B-Chat & 64.2\% & 44.0\% & 25.9\% \\
\rule{0pt}{1.5ex}
GPT3.5-Turbo-20231106$^\dag$ & 61.6\% & 43.2\% & 20.8\% & ERNIE-4-8K-0613$^\dag$ & 50.7\% & 33.8\% & 20.0\% \\
\rule{0pt}{1.5ex}
Mixtral-8x7B-Instruct & 56.5\% & 37.6\% & 19.1\% & Qwen2-7B-Instruct & 47.0\% & 26.9\% & 15.0\% \\

% ====<<<<==== Auto-generated LaTeX code end ====>>>>==== %
        \bottomrule
    \end{tabular}
\end{table}

We evaluate various LLMs across different scales and types using \Bench. 
Our analysis aims to address the following key questions concerning the system messages: 
First, at the most granular level, are LLMs capable of following each kinds of constraints? (\S\ref{subsec:csr}). Second, during individual conversation turns, how do LLMs respond to user instructions in different alignment? (\S\ref{subsec:isr}). Third, do LLMs maintain stability across multi-turn dependent or parallel conversations? (\S\ref{subsec:ssr}).
% \begin{enumerate}
% \item At the most granular level, are LLMs capable of following each kinds of constraints? (\S\ref{subsec:csr})
% % set forth in the system messages
% \item During individual conversation turns, how do LLMs respond to user instructions in different alignment? (\S\ref{subsec:isr})
% % specific (aligned) or contradictory (misaligned) 
% \item At a macro level, do LLMs maintain stability across multi-turn dependent or parallel conversations? (\S\ref{subsec:ssr})
% % What if set historical dialog to the ground truth? \S\ref{subsec:history}
% \end{enumerate}

\textbf{Metrics.} We employ the metrics outlined in \S\ref{subsec:eval_proto} to evaluate the performance of each model.
% , with each metric tailored to its corresponding level of granularity. 
Specifically, CSR reflects the model performance at the constraint granularity level and is applied in \S\ref{subsec:csr}, ISR measures the following ability of LLMs at the instruction level, as discussed in \S\ref{subsec:isr}, while SSR is utilized to assess multi-turn stability in \S\ref{subsec:ssr}.

\textbf{Settings.} We select GPT-4o as the model-based verifier in verification stage due to its demonstrated superior quality-price ratio, and set temperature to 0 to ensure deterministic output. During the generation stage, we maintain all inference parameters at their default settings across all scenarios.

\textbf{Models.} We evaluate sixteen popular LLMs including GPT family, Claude-3, Qwen-2, ERNIE-4, Moonshot, Mixtral, DeepSeek-V2, GLM-4, and Llama-3.1 family \citep{DBLP:conf/nips/BrownMRSKDNSSAA20,openai2024gpt4technicalreport,anthropic2024claude,yang2024qwen2technicalreport,sun2021ernie30largescaleknowledge,moonshot,jiang2024mixtralexperts,deepseekai2024deepseekv2strongeconomicalefficient,glm2024chatglmfamilylargelanguage,dubey2024llama3herdmodels}.
The overall results under our proposed metrics are displayed in Table \ref{tab:overall} and analyzed from the bottom up in the rest of section.
% of these models evaluated 
% layer by layer, from the most detailed aspects to the broader ones.
% \CA{, to address the questions we initially posed}

\subsection{Constraints-categorized Results}
\label{subsec:csr}

\begin{figure}[t]
\begin{minipage}{.48\linewidth}
    \centering
    \includegraphics[width=0.7\linewidth]{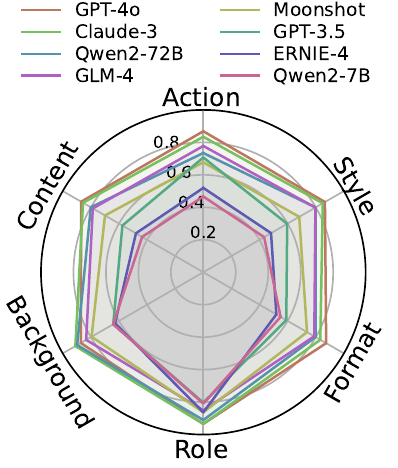}
    \captionof{figure}{The \textbf{CSR} under different types of constraints. Only 8 representative ones are shown.}
    \label{fig:radar}
\end{minipage}\hfill
\begin{minipage}{.48\linewidth}
    \centering
    \small
    \setlength{\tabcolsep}{3pt}
    \captionof{table}{The \textbf{ISR} of models (version numbers are omitted for clarity; see Table \ref{tab:overall} for details).}
    \label{tab:align}
    \begin{tabular}{@{ }c|cc|c@{ }}
        \toprule
        % \rule{0pt}{2.0ex}
        Model & Aligned & Misaligned & Total\\
        \midrule
% ====<<<<==== Auto-generated LaTeX code begin ====>>>>==== %

% --- generated by tab3_align.py --- %

\rule{0pt}{2.0ex}
GPT-4-Turbo & \underline{76.6\%} & \textbf{76.5\%} & \textbf{76.6\%} \\
\rule{0pt}{1.5ex}
GPT-4o & \textbf{77.8\%} & \underline{71.4\%} & \underline{76.4\%} \\
\rule{0pt}{1.5ex}
Claude-3 & 75.8\% & 68.3\% & 74.1\% \\
\rule{0pt}{1.5ex}
Qwen2.5-72B & 68.9\% & 56.5\% & 66.2\% \\
\rule{0pt}{1.5ex}
GLM-4 & 67.5\% & 58.5\% & 65.5\% \\
\rule{0pt}{1.5ex}
Qwen2-72B & 67.4\% & 52.6\% & 64.1\% \\
\rule{0pt}{1.5ex}
DeepSeek-V2 & 65.1\% & 49.5\% & 61.7\% \\
\rule{0pt}{1.5ex}
Llama3.1-70B & 60.7\% & 59.0\% & 60.3\% \\
\rule{0pt}{1.5ex}
Moonshot & 54.7\% & 43.7\% & 52.3\% \\
\rule{0pt}{1.5ex}
Llama3.1-8B & 48.2\% & 42.3\% & 46.9\% \\
\rule{0pt}{1.5ex}
Mixtral-8x22B & 46.0\% & 38.8\% & 44.4\% \\
\rule{0pt}{1.5ex}
GLM-4-9B & 48.3\% & 28.4\% & 44.0\% \\
\rule{0pt}{1.5ex}
GPT-3.5 & 41.9\% & 47.7\% & 43.2\% \\
\rule{0pt}{1.5ex}
Mixtral-8x7B & 39.5\% & 31.0\% & 37.6\% \\
\rule{0pt}{1.5ex}
ERNIE-4 & 37.5\% & 20.8\% & 33.8\% \\
\rule{0pt}{1.5ex}
Qwen2-7B & 29.3\% & 18.6\% & 26.9\% \\

% ====<<<<==== Auto-generated LaTeX code end ====>>>>==== %
    \bottomrule
    \end{tabular}
\end{minipage}
\end{figure}

\begin{table}[t]
    \centering
    \small
    
    \caption{The multi-turn stability results (version numbers are omitted for clarity; see Table \ref{tab:overall} for details). The $R_n$ columns indicates the percentage of the first n rounds of model responses that are all available, so the values decrease as n increases and satisfy Average($R_n$)=SSR by definition. The $k$ denotes the linear regression slope of $R_n$.}
    \label{tab:multi-turn}
    \begin{tabular}{c|c|ccccc|cc}
        \toprule
        Multi-Turn & \multirow{2}{*}{Model} & \multirow{2}{*}{R1} & \multirow{2}{*}{R2} & \multirow{2}{*}{R3} & \multirow{2}{*}{R4} & \multirow{2}{*}{R5} & \multirow{2}{*}{$-k$} &\multirow{2}{*}{\textbf{SSR}}\\
        Dependency & & & & & & & \\
        \midrule
% ====<<<<==== Auto-generated LaTeX code begin ====>>>>==== %

% --- generated by tab4_turn.py --- %

\rule{0pt}{2.0ex}
\multirow{16}{*}{Dependent} & GPT-4o & \textbf{84.8\%} & \textbf{68.5\%} & \textbf{53.1\%} & \textbf{43.3\%} & \textbf{33.7\%} & 0.128 & \textbf{56.7\%} \\
\rule{0pt}{1.5ex}
 & GPT-4-Turbo & 81.7\% & \underline{64.0\%} & 51.7\% & \underline{41.0\%} & \underline{32.3\%} & 0.122 & \underline{54.2\%} \\
\rule{0pt}{1.5ex}
 & Claude-3 & \underline{82.3\%} & \underline{64.0\%} & \underline{52.0\%} & 38.2\% & 28.4\% & 0.134 & 53.0\% \\
\rule{0pt}{1.5ex}
 & Qwen2.5-72B & 78.7\% & 58.1\% & 41.9\% & 27.8\% & 16.3\% & 0.155 & 44.6\% \\
\rule{0pt}{1.5ex}
 & DeepSeek-V2 & 77.5\% & 56.5\% & 38.2\% & 23.9\% & 12.1\% & 0.163 & 41.6\% \\
\rule{0pt}{1.5ex}
 & Qwen2-72B & 78.9\% & 54.8\% & 37.9\% & 23.3\% & 12.4\% & 0.165 & 41.5\% \\
\rule{0pt}{1.5ex}
 & GLM-4 & 76.7\% & 54.2\% & 36.2\% & 24.2\% & 15.7\% & 0.152 & 41.4\% \\
\rule{0pt}{1.5ex}
 & Llama3.1-70B & 69.9\% & 48.3\% & 32.0\% & 22.2\% & 17.7\% & 0.131 & 38.0\% \\
\rule{0pt}{1.5ex}
 & Moonshot & 69.4\% & 41.3\% & 27.2\% & 14.6\% & 8.4\% & 0.149 & 32.2\% \\
\rule{0pt}{1.5ex}
 & GLM-4-9B & 66.6\% & 38.8\% & 22.8\% & 7.9\% & 3.7\% & 0.157 & 27.9\% \\
\rule{0pt}{1.5ex}
 & Llama3.1-8B & 62.9\% & 34.3\% & 18.3\% & 9.0\% & 6.7\% & 0.138 & 26.2\% \\
\rule{0pt}{1.5ex}
 & Mixtral-8x22B & 63.5\% & 29.2\% & 13.5\% & 6.5\% & 3.7\% & 0.142 & 23.3\% \\
\rule{0pt}{1.5ex}
 & GPT-3.5 & 53.9\% & 28.7\% & 16.3\% & 7.3\% & 5.1\% & 0.119 & 22.2\% \\
\rule{0pt}{1.5ex}
 & ERNIE-4 & 61.8\% & 26.7\% & 12.6\% & 6.2\% & 2.0\% & 0.140 & 21.9\% \\
\rule{0pt}{1.5ex}
 & Mixtral-8x7B & 54.2\% & 26.4\% & 13.8\% & 6.2\% & 2.5\% & 0.124 & 20.6\% \\
\rule{0pt}{1.5ex}
 & Qwen2-7B & 52.5\% & 20.5\% & 6.5\% & 2.2\% & 1.1\% & 0.121 & 16.6\% \\
\midrule
\rule{0pt}{2.0ex}
\multirow{16}{*}{Parallel} & GPT-4o & \textbf{77.8\%} & 60.4\% & 46.5\% & 33.3\% & \underline{26.4\%} & 0.130 & \textbf{54.4\%} \\
\rule{0pt}{1.5ex}
 & GPT-4-Turbo & 72.9\% & \textbf{62.5\%} & \textbf{49.3\%} & \textbf{38.2\%} & \textbf{30.6\%} & 0.109 & \underline{53.2\%} \\
\rule{0pt}{1.5ex}
 & Claude-3 & \underline{75.7\%} & \textbf{62.5\%} & \underline{47.2\%} & \underline{34.7\%} & 25.0\% & 0.129 & 51.8\% \\
\rule{0pt}{1.5ex}
 & Qwen2.5-72B & 68.8\% & 51.4\% & 35.4\% & 22.9\% & 14.6\% & 0.137 & 42.8\% \\
\rule{0pt}{1.5ex}
 & GLM-4 & 71.5\% & 51.4\% & 40.3\% & 25.7\% & 20.8\% & 0.127 & 41.6\% \\
\rule{0pt}{1.5ex}
 & Qwen2-72B & 72.2\% & 48.6\% & 39.6\% & 28.5\% & 20.1\% & 0.124 & 41.6\% \\
\rule{0pt}{1.5ex}
 & DeepSeek-V2 & 69.4\% & 44.4\% & 30.6\% & 18.1\% & 11.1\% & 0.143 & 39.6\% \\
\rule{0pt}{1.5ex}
 & Llama3.1-70B & 66.7\% & 43.1\% & 25.7\% & 17.4\% & 11.8\% & 0.135 & 36.6\% \\
\rule{0pt}{1.5ex}
 & Moonshot & 58.3\% & 30.6\% & 17.4\% & 11.8\% & 5.6\% & 0.124 & 30.0\% \\
\rule{0pt}{1.5ex}
 & GLM-4-9B & 52.8\% & 27.8\% & 15.3\% & 5.6\% & 2.8\% & 0.122 & 25.9\% \\
\rule{0pt}{1.5ex}
 & Llama3.1-8B & 53.5\% & 25.0\% & 18.1\% & 8.3\% & 3.5\% & 0.117 & 24.9\% \\
\rule{0pt}{1.5ex}
 & Mixtral-8x22B & 56.2\% & 27.1\% & 16.7\% & 6.2\% & 2.1\% & 0.129 & 22.8\% \\
\rule{0pt}{1.5ex}
 & GPT-3.5 & 45.1\% & 18.8\% & 11.1\% & 8.3\% & 2.1\% & 0.097 & 20.8\% \\
\rule{0pt}{1.5ex}
 & ERNIE-4 & 46.5\% & 20.1\% & 7.6\% & 2.8\% & 0.7\% & 0.109 & 20.0\% \\
\rule{0pt}{1.5ex}
 & Mixtral-8x7B & 42.4\% & 20.1\% & 9.7\% & 4.9\% & 0.0\% & 0.100 & 19.1\% \\
\rule{0pt}{1.5ex}
 & Qwen2-7B & 36.1\% & 11.1\% & 6.2\% & 2.8\% & 0.0\% & 0.081 & 15.0\% \\
 
% p_value = 0.002799465792599415erq
% ====<<<<==== Auto-generated LaTeX code end ====>>>>==== %
        \bottomrule
    \end{tabular}
\end{table}

Diving into the finest granularity, we analyze the constraints embedded within each instruction and system message using the Constraint Satisfaction Rate (\textbf{CSR}) as our metric. We classify all constraints into six distinct types, as detailed in \S\ref{subsec:bench_design}, and compute the respective CSR scores. The overall CSR scores for each model are tabulated in Table \ref{tab:overall}, while the results categorized by constraint type are shown in Figure \ref{fig:radar}. We only present the most representative eight models in the radar plot, without loss of generality, and the full numeric results can be found in Appendix \ref{subsec:abs-exp}.

Overall, performance at the constraint-categorized level aligns closely with total CSR scores. GPT-4o leads the performance across all category of constraints, with Claude-3 closely behind. Qwen2-72B and GLM-4 also demonstrate strong performance.
% , while Moonshot and GPT-3.5 slightly lag behind. 
It is noteworthy that although ERNIE-4 exhibits well performance in existing instruction following benchmarks \citep{zhang2024cfbench}, its ability to follow system messages leaves room for improvement, with its performance comparable to the Qwen2-7B. 
For the better-performing models, their profiles approximate a positive hexagon in Figure \ref{fig:radar}, indicating similar CSR across each category. In contrast, models with poorer overall performance show significant variance among different types of constraints. For example, the CSR for Qwen2-7B under \textit{role constraints} is relatively high at 81.0\%, even comparable to some of the more successful models, yet it drops markedly to 43.3\% under \textit{style constraints}, highlighting an intuitive weakness. Similar patterns are also observed in GPT-3.5 and ERNIE-4.

% Given that the CSR metric
CSR is assessed at the most granular level, directly mirroring the model’s capability to satisfy constraints. 
The absolute magnitude of this value indicates the strength of the model's performance, whereas misalignment in its relative magnitude across different constraint categories highlights specific areas of weakness.
This detailed information can guide developers in enhancing model performance by focusing training efforts on these identified sub-tasks.
% \CA{Figure \ref{fig:radar} is a radar figure, a.k.a. spider figure.}

\subsection{Instruction Alignment}
\label{subsec:isr}

Moving onto the instruction level, the Instruction Satisfaction Rate (\textbf{ISR}) metric quantifies the proportion of responses generated by the models that fully follow all the given constraints. We categorize all instructions based on their alignment with the corresponding system messages, as stated in \S\ref{subsec:%  into two sets
bench_design}.
% , and calculate the ISR for each set. 
The results are displayed in Table \ref{tab:align}.

GPT-4-Turbo holds first place by a narrow margin in this instance, differing from the leader in SSR. It is observed, as expected, that performance in aligned categories generally surpasses that in misaligned categories for most models, especially GLM-4-9B. 
Compared to aligned instructions, when a user instruction conflicts with a system message (i.e., is misaligned), the model should prioritize the system message due to its higher importance. 
Such performance degradation observed in this misaligned category is likely due to the model's insufficient recognition of the system message's priority, highlighting significant potential for optimization in this area.
Although the aligned instructions do not conflict with their corresponding system messages, the ISR still informatively indicates whether or not the models satisfy the constraints embedded in both system message and user instruction. Violating any of them will result in a negative contribution. This contrasts with misaligned instructions, where satisfying the system message alone is the requirement.

It should be noted that some models exhibit minimal performance variation between the two alignment types of instructions. And surprisingly, GPT-3.5 get notably higher ISR score on misaligned instructions than the aligned set, suggesting its acute awareness of the prioritization required by the system message.
Furthermore, this performance hints at the capability of LLMs to effectively prioritize system messages when faced with contradictory instructions.
% , potentially meeting or even exceeding their ability to follow self-consistency constraints.

\subsection{Multi-turn Stability}
\label{subsec:ssr}
% \CA{$-k$: p-value is 0.003}

On a broader scale, we categorize all conversations into multi-turn dependent and multi-turn parallel, as mentioned in \S\ref{subsec:bench_design}. To evaluate the stability maintenance capability of large language models across multi-turn conversations, we utilize the Session Stability Rate (\textbf{SSR}) as the metric. Additionally, we report the $R_n$ values, defining as the percentage of all sessions in which the first n rounds of model responses follow all given constraints. The linear regression slopes of $R_n$ (denoted as $k$) are also reported in Table \ref{tab:multi-turn}, illustrating the degradation in model performance across successive conversation turns.

The model GPT-4o and GPT-4-Turbo outperform other models in terms of SSR, while Claude-3 follows closely behind. 
From the analysis of the slopes, we observe the $R_n$ values of multi-turn dependent conversations generally decline more rapidly than those of parallel ones.
This difference may be attribute to the simpler and more straightforward instruction in the first turn of multi-turn dependent conversation,
while subsequent rounds involve less irrelevant contextual references, which has been well-studied in prior works such as \citep{DBLP:conf/icml/ShiCMSDCSZ23}. % cite: Large Language Models Can Be Easily Distracted by Irrelevant Context
Interestingly, some models like GPT-4-turbo do not exhibit significant differentiation between these two categories of conversation,
% , probably implying that their architecture or training may equip them with a robust ability to concentrate on more valuable context.
while some other models such as ERINE-4 and Claude-3, perform moderately well in the first round (with high $R_1$ value), but degraded more rapidly in subsequent dialogues, as evidenced by $R_n$ values decaying at a higher rate, resulting in lower SSR. 
This observation highlights a potential area for improvement in the ability of some models to process multi-turn conversations or manage long contexts within system message constraints.
%, enlightening developers to optimize their models in the specified direction.

% Moreover, we find that some models, such as ERINE-4 and Claude-3, perform moderately well in the first round (with high $R_1$ value), but the performance of subsequent dialogues degraded more rapidly, as evidenced by $R_n$ values decaying at a higher rate, resulting in lower SSR.  This observation highlights a potential area for improvement in the ability of some models to process multi-turn conversations or manage long contexts within system message constraints, enlightening developers to optimize their models in the specified direction.

Overall, the SSR metric measures models from a high-level perspective, focusing on stability across multiple conversation turns. 
% With multi-turn conversations categorized based on their dependency, 
The reported results show the performance differences and reveal variations in each model's ability to maintain stable over multiple rounds with system message. This provides developers with a macro-view reference of each model's capacity, 
%to follow and process the system messages, 
guiding further development and optimization efforts to enhance their performance in complex conversational scenarios, since the best SSR is only 54.4\%.

% \YZ{TODO: We may put more emphasis on the fact that we are in system message scenario in \S 4.2 to 4.4.}

% \CA{ ----- draft : points to state -----
% \begin{itemize}
% \item Rn decreases by definition.
% \item GPT-4(s) outperform others.
% \item Dependent performs better than Parallel (why?).
% \item Some models like GPT-4-turbo shows little difference in performance between the two categories.
% \item Some models like ERNIE-4 getting much faster downgrades as the conversation turn.
% \end{itemize}
% }

% \subsection{Historical Dialogue and Multi-Turn Stability}
\subsection{Investigative Experiments}
\label{subsec:history}

% \begin{itemize}
%     \item We observe a strong correlation between the ability of models to follow system messages and the proportion of the attention score attributed to their tokens. 
%     \begin{itemize}
%         \item On one hand, we found that the trend of changes in attention scores in multi-turn dialogues is consistent with the experimental results (slope parallel vs dependent; trend of llama vs glm).
%         \item On the other hand, after replacing the content of the system message with the user message,
%     \end{itemize}
%     \item We also interestingly find some peaks at the beginnings of each turn, implying that the inherent relationship between the system message and incoming user tokens has been detected by models. This phenomenon may provide a perspective for explaining why the system message can continuously has a effect on multi-turn conversations.
% \end{itemize}

\begin{figure}[t]
    \centering
    \includegraphics[width=0.98\linewidth]{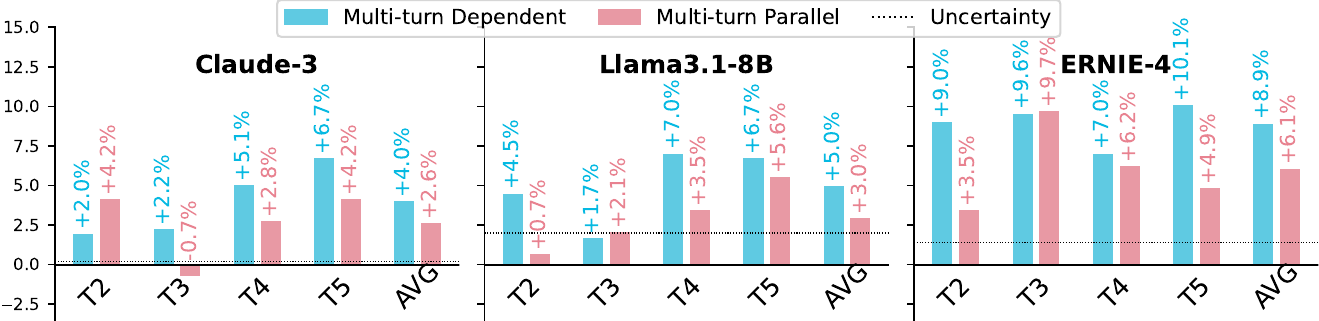}
    \caption{The ISR gain when using ground-truth history; the n-th turn is denoted as $T_n$.}
    \label{fig:history-gt}
\end{figure}

\paragraph{How Does Historical Dialogue Affect the Multi-turn Stability?} 
% \paragraph{How Historical Dialogue Affects the Multi-turn Stability?} 
We further explore the effects of history conversations, since it might be one of the potential factors for multi-turn stability.
% In order to explore the above hypothesis, 
We replace the historical model response with the ground truth, comparing the ISR improvement of each turn throughout a session. The results are shown in Figure \ref{fig:history-gt}.
Overall, the correctness of historical response has a positive impact on the performance of the model. Besides, the improvement for multi-turn dependent conversations is more apparent than the parallel ones. In the case of multi-turn parallel, the magnitude of changes is comparable to the random oscillations in the first round (i.e., $|\Delta T_1|$, plotted as the "Uncertainty" lines in Figure \ref{fig:history-gt}). Among the presented models, ERINE-4 has the sharpest decline with round increasing in Table \ref{tab:multi-turn}, but its improvement is the most obvious with correct history dialogue, suggesting that developers need to pay more attention to the historical errors in multi-round conversations.
% in each round 

%\paragraph{How to Explain System Instruction Following from the Perspective of Attention Scores?} 

% \CA{Note: We should state what and how Figure \ref{fig:atten-score} is plotted in either caption or main text. I prefer main text. (Average across all heads, all layers, and all system messages for (a)(c), and one arbitrarily chosen system message for (b).)}

\begin{figure}[t]
    \centering
    \includegraphics[width=1.0\linewidth]{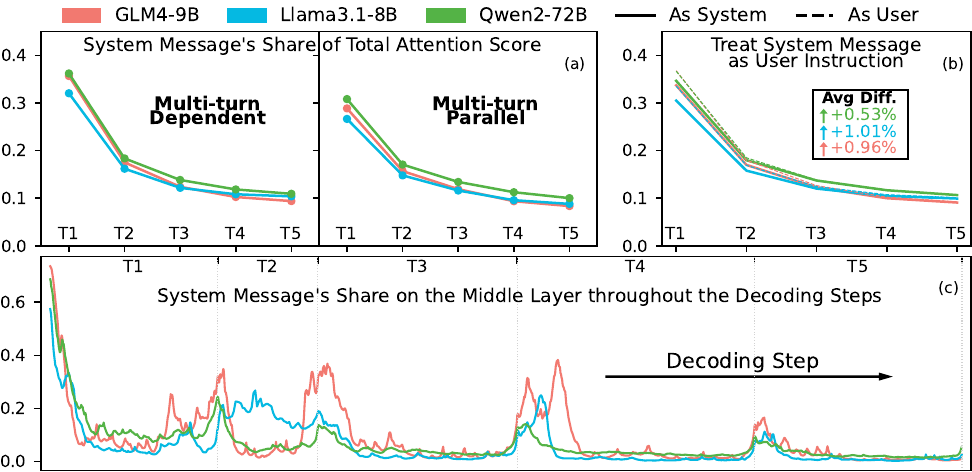}
    \caption{Proportion of total attention score attributed to system message.  (a)(b) The average ratios for each turn, calculated across all heads and layers throughout specified categories of session (or entire dataset); the dashed lines in (b) indicates scenarios where the role of system message is replaced with ``user''. (c) The ratio for each token throughout a whole session, containing 5 turns.}
    \label{fig:atten-score}
\end{figure}

\paragraph{Is There a Correlation Between Attention Distribution and the Ability to Follow System Messages?} 
% \paragraph{Correlation Between Attention Distribution and System Messages Following Ability?} 
We observe a strong correlation between the ability of models to follow system messages and the proportion of the attention score attributed to their tokens. 
To illustrate our findings, we select three open-source models (GLM-4-9B, Llama3.1-8B, and Qwen2-72B) for analysis. The solid lines in Figure \appref{fig:atten-score}{a} and \appref{fig:atten-score}{b} indicates the average proportion of attention scores attributed to system messages, calculated across all heads and layers throughout the corresponding system messages set. The attention score proportion of Llama3.1-8B is lower than that of GLM-4-9B in the first three turns, and vice versa afterwards, which is consistent with the relative performance of the two models on $R_n$ in Table \ref{tab:multi-turn}. Besides, the decline trend of multi-turn dependent instances is steeper than multi-turn parallel, which matches the difference in the slope k of the two sets of data. 
We also conduct case study to explore further at token granularity, and arbitrarily choose one session to present in Figure \appref{fig:atten-score}{c}. 
We interestingly find some peaks at the beginnings of each turn, implying that the inherent relationship between the system message and incoming user tokens has been detected by models. This phenomenon may provide a perspective for explaining why the system message can continuously has a effect on multi-turn conversations.
% we find that the trend of attention score changes is consistent with the experimental results .
%Figure \appref{fig:atten-score}{a} displays the attention ratio at the middle layer for each token throughout the session. The x-axis is scaled to align with each conversation turn for ease of comparison, accommodating the varying output lengths among the models.
%Overall, models with higher ISR scores tend to focus more on the system message tokens, as evidenced by the higher average ratios shown in Figure \appref{fig:atten-score}{a}, and vice versa.
% Conversely, models with lower ISR show less attention to these tokens. 
%We also interestingly find some peaks at the beginnings of each turn, implying that the inherent relationship between the system message and incoming user tokens has been detected by models, and those with higher peak generally achieve better scores.

%in \Bench. 
% \subsection{Case Study: Investigating Attention Score}

\paragraph{Are System Messages and User Messages Truly Treated Differently?} 
% \paragraph{Clear Distinction Between System Message and User Message?} 
In the inference stage, the only difference between system messages and user messages is the different marker tokens at the beginning of the text (e.g., \texttt{<system>}, \texttt{<user>}). To investigate whether the model exhibits different levels of attention due to the marker token when processing system messages and user messages, we repurpose the text originally designated as a system message to serve as a user message, and collected the attention scores attributed to the same text under these two distinct scenarios. As shown in Figure \appref{fig:atten-score}{b}, the changes in attention score before and after replacing the text of system messages with user messages are very weak. This indicates that there is no strict distinction between system messages and user messages during the inference stage, and the capability of following system message is more influenced by the construction strategy of the training data.
%To investigate whether the model exhibits different levels of attention when processing system messages compared to user prompts, we stick on the selected session dialogue in the last paragraph, repurposing the text originally designated as system message to serve as user prompt, and collects the attention scores attributed to the same text under these two distinct scenarios. 
%We hypothesize that there would be a decrease in attention scores when the text is perceived as user input, reflecting a lower prioritization by the models.
%The average percentages of attention scores held by such text segment are presented in Figure \appref{fig:atten-score}{b}. 
%Qwen2-72B exhibits a notable decline in attention ratio, aligning with our expectations, while Llama3.1-8B also shows a slight decline.
%However, GLM-4-9B presents an opposite trend, with an increase in attention ratio, indicating its lack of prioritization in processing system messages. 
%This characteristic is also reflected in its poorer performance on the misaligned instructions set (Table \ref{tab:align}), since it can be easily distracted by contradict user instructions without adequately prioritizing system messages, and shows room for further improvement.

% \CA{Table \ref{tab:multi-bench} corroborates the distinctive nature of our bench in comparison to existing benchmarks.}
\section{Conclusion}
We propose \Bench, the first comprehensive benchmark evaluating the system message following ability of large language models. 
\Bench constructs system messages and corresponding user instructions based on six types of well-designed constraints, differentiates between aligned and misaligned instructions at the instruction level, and categorizes multi-turn conversations based on their dependency. It is consist of 500 sessions with a total of 2500 turns of high-quality conversations. 
Additionally, \Bench also proposes three-level granularity metrics to comprehensively measure the model performance in terms of constraint-level following, instruction-level satisfaction, and multi-turn stability.
Our experiments across various large language models demonstrate significant differentiation in model scores under \Bench in multiple perspectives. 
% These results highlight \Bench's effectiveness in performance assessment and its utility in improving and enhancing modern large language models.
These results not only underscore the effectiveness of \Bench in performance assessment but also offer valuable insights for model improvement, confirming \Bench's utility. 
% This confirms \Bench's utility in enhancing the performance and utility of modern large language models.

% AAAI creates proceedings, working notes, and technical reports directly from electronic source furnished by the authors. To ensure that all papers in the publication have a uniform appearance, authors must adhere to the following instructions.

\bibliography{iclr2025_conference}

\begin{thebibliography}{38}
\providecommand{\natexlab}[1]{#1}
\providecommand{\url}[1]{\texttt{#1}}
\expandafter\ifx\csname urlstyle\endcsname\relax
  \providecommand{\doi}[1]{doi: #1}\else
  \providecommand{\doi}{doi: \begingroup \urlstyle{rm}\Url}\fi

\bibitem[AlKhamissi et~al.(2024)AlKhamissi, ElNokrashy, AlKhamissi, and Diab]{mistrial}
Badr AlKhamissi, Muhammad~N. ElNokrashy, Mai AlKhamissi, and Mona~T. Diab.
\newblock Investigating cultural alignment of large language models.
\newblock \emph{CoRR}, abs/2402.13231, 2024.
\newblock \doi{10.48550/ARXIV.2402.13231}.
\newblock URL \url{https://doi.org/10.48550/arXiv.2402.13231}.

\bibitem[Anthropic(2024)]{anthropic2024claude}
Anthropic.
\newblock The claude 3 model family: Opus, sonnet, haiku.
\newblock \url{https://www-cdn.anthropic.com/de8ba9b01c9ab7cbabf5c33b80b7bbc618857627/Model_Card_Claude_3.pdf}, 2024.
\newblock "Accessed: 2024-08-14".

\bibitem[Brown et~al.(2020)Brown, Mann, Ryder, Subbiah, Kaplan, Dhariwal, Neelakantan, Shyam, Sastry, Askell, Agarwal, Herbert{-}Voss, Krueger, Henighan, Child, Ramesh, Ziegler, Wu, Winter, Hesse, Chen, Sigler, Litwin, Gray, Chess, Clark, Berner, McCandlish, Radford, Sutskever, and Amodei]{DBLP:conf/nips/BrownMRSKDNSSAA20}
Tom~B. Brown, Benjamin Mann, Nick Ryder, Melanie Subbiah, Jared Kaplan, Prafulla Dhariwal, Arvind Neelakantan, Pranav Shyam, Girish Sastry, Amanda Askell, Sandhini Agarwal, Ariel Herbert{-}Voss, Gretchen Krueger, Tom Henighan, Rewon Child, Aditya Ramesh, Daniel~M. Ziegler, Jeffrey Wu, Clemens Winter, Christopher Hesse, Mark Chen, Eric Sigler, Mateusz Litwin, Scott Gray, Benjamin Chess, Jack Clark, Christopher Berner, Sam McCandlish, Alec Radford, Ilya Sutskever, and Dario Amodei.
\newblock Language models are few-shot learners.
\newblock In Hugo Larochelle, Marc'Aurelio Ranzato, Raia Hadsell, Maria{-}Florina Balcan, and Hsuan{-}Tien Lin (eds.), \emph{Advances in Neural Information Processing Systems 33: Annual Conference on Neural Information Processing Systems 2020, NeurIPS 2020, December 6-12, 2020, virtual}, 2020.
\newblock URL \url{https://proceedings.neurips.cc/paper/2020/hash/1457c0d6bfcb4967418bfb8ac142f64a-Abstract.html}.

\bibitem[DeepSeek-AI(2024)]{deepseekai2024deepseekv2strongeconomicalefficient}
DeepSeek-AI.
\newblock Deepseek-v2: A strong, economical, and efficient mixture-of-experts language model, 2024.
\newblock URL \url{https://arxiv.org/abs/2405.04434}.

\bibitem[He et~al.(2024)He, Zeng, Huang, Chen, Xiao, He, Zhou, Liang, and Xiao]{CELLO}
Qianyu He, Jie Zeng, Wenhao Huang, Lina Chen, Jin Xiao, Qianxi He, Xunzhe Zhou, Jiaqing Liang, and Yanghua Xiao.
\newblock Can large language models understand real-world complex instructions?
\newblock In Michael~J. Wooldridge, Jennifer~G. Dy, and Sriraam Natarajan (eds.), \emph{Thirty-Eighth {AAAI} Conference on Artificial Intelligence, {AAAI} 2024, Thirty-Sixth Conference on Innovative Applications of Artificial Intelligence, {IAAI} 2024, Fourteenth Symposium on Educational Advances in Artificial Intelligence, {EAAI} 2014, February 20-27, 2024, Vancouver, Canada}, pp.\  18188--18196. {AAAI} Press, 2024.
\newblock \doi{10.1609/AAAI.V38I16.29777}.
\newblock URL \url{https://doi.org/10.1609/aaai.v38i16.29777}.

\bibitem[Jiang et~al.(2024)Jiang, Sablayrolles, Roux, Mensch, Savary, Bamford, Chaplot, de~las Casas, Hanna, Bressand, Lengyel, Bour, Lample, Lavaud, Saulnier, Lachaux, Stock, Subramanian, Yang, Antoniak, Scao, Gervet, Lavril, Wang, Lacroix, and Sayed]{jiang2024mixtralexperts}
Albert~Q. Jiang, Alexandre Sablayrolles, Antoine Roux, Arthur Mensch, Blanche Savary, Chris Bamford, Devendra~Singh Chaplot, Diego de~las Casas, Emma~Bou Hanna, Florian Bressand, Gianna Lengyel, Guillaume Bour, Guillaume Lample, Lélio~Renard Lavaud, Lucile Saulnier, Marie-Anne Lachaux, Pierre Stock, Sandeep Subramanian, Sophia Yang, Szymon Antoniak, Teven~Le Scao, Théophile Gervet, Thibaut Lavril, Thomas Wang, Timothée Lacroix, and William~El Sayed.
\newblock Mixtral of experts, 2024.
\newblock URL \url{https://arxiv.org/abs/2401.04088}.

\bibitem[Jiang et~al.(2023)Jiang, Wang, Zeng, Zhong, Li, Mi, Shang, Jiang, Liu, and Wang]{followbench}
Yuxin Jiang, Yufei Wang, Xingshan Zeng, Wanjun Zhong, Liangyou Li, Fei Mi, Lifeng Shang, Xin Jiang, Qun Liu, and Wei Wang.
\newblock Followbench: {A} multi-level fine-grained constraints following benchmark for large language models.
\newblock \emph{CoRR}, abs/2310.20410, 2023.
\newblock \doi{10.48550/ARXIV.2310.20410}.
\newblock URL \url{https://doi.org/10.48550/arXiv.2310.20410}.

\bibitem[Lee et~al.(2024)Lee, Park, Kim, and Seo]{align_preference}
Seongyun Lee, Sue~Hyun Park, Seungone Kim, and Minjoon Seo.
\newblock Aligning to thousands of preferences via system message generalization.
\newblock \emph{CoRR}, abs/2405.17977, 2024.
\newblock \doi{10.48550/ARXIV.2405.17977}.
\newblock URL \url{https://doi.org/10.48550/arXiv.2405.17977}.

\bibitem[Li et~al.(2024{\natexlab{a}})Li, Liu, Bashkansky, Bau, Vi{\'e}gas, Pfister, and Wattenberg]{stability}
Kenneth Li, Tianle Liu, Naomi Bashkansky, David Bau, Fernanda Vi{\'e}gas, Hanspeter Pfister, and Martin Wattenberg.
\newblock Measuring and controlling instruction (in)stability in language model dialogs.
\newblock In \emph{First Conference on Language Modeling}, 2024{\natexlab{a}}.
\newblock URL \url{https://openreview.net/forum?id=60a1SAtH4e}.

\bibitem[Li et~al.(2024{\natexlab{b}})Li, Yan, Wang, Tang, Ren, Srinivasan, and Jin]{verbalizer}
Shiyang Li, Jun Yan, Hai Wang, Zheng Tang, Xiang Ren, Vijay Srinivasan, and Hongxia Jin.
\newblock Instruction-following evaluation through verbalizer manipulation.
\newblock In Kevin Duh, Helena G{\'{o}}mez{-}Adorno, and Steven Bethard (eds.), \emph{Findings of the Association for Computational Linguistics: {NAACL} 2024, Mexico City, Mexico, June 16-21, 2024}, pp.\  3678--3692. Association for Computational Linguistics, 2024{\natexlab{b}}.
\newblock \doi{10.18653/V1/2024.FINDINGS-NAACL.233}.
\newblock URL \url{https://doi.org/10.18653/v1/2024.findings-naacl.233}.

\bibitem[Liu et~al.(2024)Liu, Yu, Zhang, Xu, Lei, Lai, Gu, Ding, Men, Yang, Zhang, Deng, Zeng, Du, Zhang, Shen, Zhang, Su, Sun, Huang, Dong, and Tang]{AgentBench}
Xiao Liu, Hao Yu, Hanchen Zhang, Yifan Xu, Xuanyu Lei, Hanyu Lai, Yu~Gu, Hangliang Ding, Kaiwen Men, Kejuan Yang, Shudan Zhang, Xiang Deng, Aohan Zeng, Zhengxiao Du, Chenhui Zhang, Sheng Shen, Tianjun Zhang, Yu~Su, Huan Sun, Minlie Huang, Yuxiao Dong, and Jie Tang.
\newblock Agentbench: Evaluating llms as agents.
\newblock In \emph{The Twelfth International Conference on Learning Representations, {ICLR} 2024, Vienna, Austria, May 7-11, 2024}. OpenReview.net, 2024.
\newblock URL \url{https://openreview.net/forum?id=zAdUB0aCTQ}.

\bibitem[{Llama Team}(2024)]{dubey2024llama3herdmodels}
{Llama Team}.
\newblock The llama 3 herd of models, 2024.
\newblock URL \url{https://arxiv.org/abs/2407.21783}.

\bibitem[Lu et~al.(2024)Lu, Yu, Lu, Lin, Yu, Sun, Han, and Li]{SOFA}
Xinyu Lu, Bowen Yu, Yaojie Lu, Hongyu Lin, Haiyang Yu, Le~Sun, Xianpei Han, and Yongbin Li.
\newblock Sofa: Shielded on-the-fly alignment via priority rule following.
\newblock \emph{CoRR}, abs/2402.17358, 2024.
\newblock \doi{10.48550/ARXIV.2402.17358}.
\newblock URL \url{https://doi.org/10.48550/arXiv.2402.17358}.

\bibitem[Ma et~al.(2024)Ma, Mishra, Liu, Su, Chen, Kulkarni, Cheng, Le, and Chi]{beyondchatbot}
Xiao Ma, Swaroop Mishra, Ariel Liu, Sophie~Ying Su, Jilin Chen, Chinmay Kulkarni, Heng{-}Tze Cheng, Quoc~V. Le, and Ed~H. Chi.
\newblock Beyond chatbots: Explorellm for structured thoughts and personalized model responses.
\newblock In Florian~'Floyd' Mueller, Penny Kyburz, Julie~R. Williamson, and Corina Sas (eds.), \emph{Extended Abstracts of the {CHI} Conference on Human Factors in Computing Systems, {CHI} {EA} 2024, Honolulu, HI, USA, May 11-16, 2024}, pp.\  56:1--56:12. {ACM}, 2024.
\newblock \doi{10.1145/3613905.3651093}.
\newblock URL \url{https://doi.org/10.1145/3613905.3651093}.

\bibitem[{Moonshot AI}(2023)]{moonshot}
{Moonshot AI}.
\newblock Moonshot website.
\newblock \url{https://platform.moonshot.cn/}, 2023.
\newblock "Accessed: 2024-08-14".

\bibitem[Mu et~al.(2023)Mu, Chen, Wang, Chen, Karamardian, Aljeraisy, Hendrycks, and Wagner]{can_rules}
Norman Mu, Sarah Chen, Zifan Wang, Sizhe Chen, David Karamardian, Lulwa Aljeraisy, Dan Hendrycks, and David~A. Wagner.
\newblock Can llms follow simple rules?
\newblock \emph{CoRR}, abs/2311.04235, 2023.
\newblock \doi{10.48550/ARXIV.2311.04235}.
\newblock URL \url{https://doi.org/10.48550/arXiv.2311.04235}.

\bibitem[Mukherjee et~al.(2023)Mukherjee, Mitra, Jawahar, Agarwal, Palangi, and Awadallah]{orca}
Subhabrata Mukherjee, Arindam Mitra, Ganesh Jawahar, Sahaj Agarwal, Hamid Palangi, and Ahmed Awadallah.
\newblock Orca: Progressive learning from complex explanation traces of {GPT-4}.
\newblock \emph{CoRR}, abs/2306.02707, 2023.
\newblock \doi{10.48550/ARXIV.2306.02707}.
\newblock URL \url{https://doi.org/10.48550/arXiv.2306.02707}.

\bibitem[Nakano et~al.(2021)Nakano, Hilton, Balaji, Wu, Ouyang, Kim, Hesse, Jain, Kosaraju, Saunders, Jiang, Cobbe, Eloundou, Krueger, Button, Knight, Chess, and Schulman]{WebGPT}
Reiichiro Nakano, Jacob Hilton, Suchir Balaji, Jeff Wu, Long Ouyang, Christina Kim, Christopher Hesse, Shantanu Jain, Vineet Kosaraju, William Saunders, Xu~Jiang, Karl Cobbe, Tyna Eloundou, Gretchen Krueger, Kevin Button, Matthew Knight, Benjamin Chess, and John Schulman.
\newblock Webgpt: Browser-assisted question-answering with human feedback.
\newblock \emph{CoRR}, abs/2112.09332, 2021.
\newblock URL \url{https://arxiv.org/abs/2112.09332}.

\bibitem[OpenAI(2022)]{OpenAI2022}
OpenAI.
\newblock Chatgpt: Optimizing language models for dialogue, 2022.
\newblock URL \url{https://openai.com/blog/chatgpt/}.

\bibitem[OpenAI(2024)]{openai2024gpt4technicalreport}
OpenAI.
\newblock Gpt-4 technical report, 2024.
\newblock URL \url{https://arxiv.org/abs/2303.08774}.

\bibitem[Parisi et~al.(2022)Parisi, Zhao, and Fiedel]{TALM}
Aaron Parisi, Yao Zhao, and Noah Fiedel.
\newblock {TALM:} tool augmented language models.
\newblock \emph{CoRR}, abs/2205.12255, 2022.
\newblock \doi{10.48550/ARXIV.2205.12255}.
\newblock URL \url{https://doi.org/10.48550/arXiv.2205.12255}.

\bibitem[Qin et~al.(2024)Qin, Song, Hu, Yao, Cho, Wang, Wu, Liu, Liu, and Yu]{infobenchbib}
Yiwei Qin, Kaiqiang Song, Yebowen Hu, Wenlin Yao, Sangwoo Cho, Xiaoyang Wang, Xuansheng Wu, Fei Liu, Pengfei Liu, and Dong Yu.
\newblock Infobench: Evaluating instruction following ability in large language models.
\newblock \emph{CoRR}, abs/2401.03601, 2024.
\newblock \doi{10.48550/ARXIV.2401.03601}.
\newblock URL \url{https://doi.org/10.48550/arXiv.2401.03601}.

\bibitem[Ramlochan(2024)]{system_message}
Sunil Ramlochan.
\newblock System prompts in large language models.
\newblock \url{https://promptengineering.org/system-prompts-in-large-language-models/}, 2024.
\newblock "Accessed: 2024-08-16".

\bibitem[Salewski et~al.(2023)Salewski, Alaniz, Rio{-}Torto, Schulz, and Akata]{incontext}
Leonard Salewski, Stephan Alaniz, Isabel Rio{-}Torto, Eric Schulz, and Zeynep Akata.
\newblock In-context impersonation reveals large language models' strengths and biases.
\newblock In Alice Oh, Tristan Naumann, Amir Globerson, Kate Saenko, Moritz Hardt, and Sergey Levine (eds.), \emph{Advances in Neural Information Processing Systems 36: Annual Conference on Neural Information Processing Systems 2023, NeurIPS 2023, New Orleans, LA, USA, December 10 - 16, 2023}, 2023.
\newblock URL \url{http://papers.nips.cc/paper\_files/paper/2023/hash/e3fe7b34ba4f378df39cb12a97193f41-Abstract-Conference.html}.

\bibitem[Schick et~al.(2023)Schick, Dwivedi{-}Yu, Dess{\`{\i}}, Raileanu, Lomeli, Hambro, Zettlemoyer, Cancedda, and Scialom]{Toolformer}
Timo Schick, Jane Dwivedi{-}Yu, Roberto Dess{\`{\i}}, Roberta Raileanu, Maria Lomeli, Eric Hambro, Luke Zettlemoyer, Nicola Cancedda, and Thomas Scialom.
\newblock Toolformer: Language models can teach themselves to use tools.
\newblock In Alice Oh, Tristan Naumann, Amir Globerson, Kate Saenko, Moritz Hardt, and Sergey Levine (eds.), \emph{Advances in Neural Information Processing Systems 36: Annual Conference on Neural Information Processing Systems 2023, NeurIPS 2023, New Orleans, LA, USA, December 10 - 16, 2023}, 2023.
\newblock URL \url{http://papers.nips.cc/paper\_files/paper/2023/hash/d842425e4bf79ba039352da0f658a906-Abstract-Conference.html}.

\bibitem[Shi et~al.(2023)Shi, Chen, Misra, Scales, Dohan, Chi, Sch{\"{a}}rli, and Zhou]{DBLP:conf/icml/ShiCMSDCSZ23}
Freda Shi, Xinyun Chen, Kanishka Misra, Nathan Scales, David Dohan, Ed~H. Chi, Nathanael Sch{\"{a}}rli, and Denny Zhou.
\newblock Large language models can be easily distracted by irrelevant context.
\newblock In Andreas Krause, Emma Brunskill, Kyunghyun Cho, Barbara Engelhardt, Sivan Sabato, and Jonathan Scarlett (eds.), \emph{International Conference on Machine Learning, {ICML} 2023, 23-29 July 2023, Honolulu, Hawaii, {USA}}, volume 202 of \emph{Proceedings of Machine Learning Research}, pp.\  31210--31227. {PMLR}, 2023.
\newblock URL \url{https://proceedings.mlr.press/v202/shi23a.html}.

\bibitem[Sun et~al.(2021)Sun, Wang, Feng, Ding, Pang, Shang, Liu, Chen, Zhao, Lu, Liu, Wu, Gong, Liang, Shang, Sun, Liu, Ouyang, Yu, Tian, Wu, and Wang]{sun2021ernie30largescaleknowledge}
Yu~Sun, Shuohuan Wang, Shikun Feng, Siyu Ding, Chao Pang, Junyuan Shang, Jiaxiang Liu, Xuyi Chen, Yanbin Zhao, Yuxiang Lu, Weixin Liu, Zhihua Wu, Weibao Gong, Jianzhong Liang, Zhizhou Shang, Peng Sun, Wei Liu, Xuan Ouyang, Dianhai Yu, Hao Tian, Hua Wu, and Haifeng Wang.
\newblock Ernie 3.0: Large-scale knowledge enhanced pre-training for language understanding and generation, 2021.
\newblock URL \url{https://arxiv.org/abs/2107.02137}.
\newblock "Accessed: 2024-08-14".

\bibitem[{Team GLM}(2024)]{glm2024chatglmfamilylargelanguage}
{Team GLM}.
\newblock Chatglm: A family of large language models from glm-130b to glm-4 all tools, 2024.
\newblock URL \url{https://arxiv.org/abs/2406.12793}.

\bibitem[Touvron et~al.(2023)Touvron, Martin, Stone, Albert, Almahairi, Babaei, Bashlykov, Batra, Bhargava, Bhosale, Bikel, Blecher, Canton{-}Ferrer, Chen, Cucurull, Esiobu, Fernandes, Fu, Fu, Fuller, Gao, Goswami, Goyal, Hartshorn, Hosseini, Hou, Inan, Kardas, Kerkez, Khabsa, Kloumann, Korenev, Koura, Lachaux, Lavril, Lee, Liskovich, Lu, Mao, Martinet, Mihaylov, Mishra, Molybog, Nie, Poulton, Reizenstein, Rungta, Saladi, Schelten, Silva, Smith, Subramanian, Tan, Tang, Taylor, Williams, Kuan, Xu, Yan, Zarov, Zhang, Fan, Kambadur, Narang, Rodriguez, Stojnic, Edunov, and Scialom]{llama-2}
Hugo Touvron, Louis Martin, Kevin Stone, Peter Albert, Amjad Almahairi, Yasmine Babaei, Nikolay Bashlykov, Soumya Batra, Prajjwal Bhargava, Shruti Bhosale, Dan Bikel, Lukas Blecher, Cristian Canton{-}Ferrer, Moya Chen, Guillem Cucurull, David Esiobu, Jude Fernandes, Jeremy Fu, Wenyin Fu, Brian Fuller, Cynthia Gao, Vedanuj Goswami, Naman Goyal, Anthony Hartshorn, Saghar Hosseini, Rui Hou, Hakan Inan, Marcin Kardas, Viktor Kerkez, Madian Khabsa, Isabel Kloumann, Artem Korenev, Punit~Singh Koura, Marie{-}Anne Lachaux, Thibaut Lavril, Jenya Lee, Diana Liskovich, Yinghai Lu, Yuning Mao, Xavier Martinet, Todor Mihaylov, Pushkar Mishra, Igor Molybog, Yixin Nie, Andrew Poulton, Jeremy Reizenstein, Rashi Rungta, Kalyan Saladi, Alan Schelten, Ruan Silva, Eric~Michael Smith, Ranjan Subramanian, Xiaoqing~Ellen Tan, Binh Tang, Ross Taylor, Adina Williams, Jian~Xiang Kuan, Puxin Xu, Zheng Yan, Iliyan Zarov, Yuchen Zhang, Angela Fan, Melanie Kambadur, Sharan Narang, Aur{\'{e}}lien Rodriguez, Robert Stojnic, Sergey Edunov,
  and Thomas Scialom.
\newblock Llama 2: Open foundation and fine-tuned chat models.
\newblock \emph{CoRR}, abs/2307.09288, 2023.
\newblock \doi{10.48550/ARXIV.2307.09288}.
\newblock URL \url{https://doi.org/10.48550/arXiv.2307.09288}.

\bibitem[Wallace et~al.(2024)Wallace, Xiao, Leike, Weng, Heidecke, and Beutel]{hierarchy}
Eric Wallace, Kai Xiao, Reimar Leike, Lilian Weng, Johannes Heidecke, and Alex Beutel.
\newblock The instruction hierarchy: Training llms to prioritize privileged instructions.
\newblock \emph{CoRR}, abs/2404.13208, 2024.
\newblock \doi{10.48550/ARXIV.2404.13208}.
\newblock URL \url{https://doi.org/10.48550/arXiv.2404.13208}.

\bibitem[Wang et~al.(2024)Wang, Bukharin, Jiang, Yin, Wang, Zhao, Shang, Zhang, Yin, Li, Chen, and Li]{RNR}
Kuan Wang, Alexander Bukharin, Haoming Jiang, Qingyu Yin, Zhengyang Wang, Tuo Zhao, Jingbo Shang, Chao Zhang, Bing Yin, Xian Li, Jianshu Chen, and Shiyang Li.
\newblock Rnr: Teaching large language models to follow roles and rules, 2024.
\newblock URL \url{https://arxiv.org/abs/2409.13733}.

\bibitem[Wen et~al.(2024)Wen, Ke, Gu, Wu, Huang, Zhou, Li, Hu, Gao, Xu, Liu, Tang, Wang, and Huang]{constraints-composition}
Bosi Wen, Pei Ke, Xiaotao Gu, Lindong Wu, Hao Huang, Jinfeng Zhou, Wenchuang Li, Binxin Hu, Wendy Gao, Jiaxin Xu, Yiming Liu, Jie Tang, Hongning Wang, and Minlie Huang.
\newblock Benchmarking complex instruction-following with multiple constraints composition.
\newblock \emph{CoRR}, abs/2407.03978, 2024.
\newblock \doi{10.48550/ARXIV.2407.03978}.
\newblock URL \url{https://doi.org/10.48550/arXiv.2407.03978}.

\bibitem[Xia et~al.(2024)Xia, Xing, Du, Yang, Feng, Xu, Yin, and Xiong]{FIFO}
Congying Xia, Chen Xing, Jiangshu Du, Xinyi Yang, Yihao Feng, Ran Xu, Wenpeng Yin, and Caiming Xiong.
\newblock {FOFO:} {A} benchmark to evaluate llms' format-following capability.
\newblock \emph{CoRR}, abs/2402.18667, 2024.
\newblock \doi{10.48550/ARXIV.2402.18667}.
\newblock URL \url{https://doi.org/10.48550/arXiv.2402.18667}.

\bibitem[Yang et~al.(2024)Yang, Yang, Hui, Zheng, Yu, Zhou, Li, Li, Liu, Huang, Dong, Wei, Lin, Tang, Wang, Yang, Tu, Zhang, Ma, Yang, Xu, Zhou, Bai, He, Lin, Dang, Lu, Chen, Yang, Li, Xue, Ni, Zhang, Wang, Peng, Men, Gao, Lin, Wang, Bai, Tan, Zhu, Li, Liu, Ge, Deng, Zhou, Ren, Zhang, Wei, Ren, Liu, Fan, Yao, Zhang, Wan, Chu, Liu, Cui, Zhang, Guo, and Fan]{yang2024qwen2technicalreport}
An~Yang, Baosong Yang, Binyuan Hui, Bo~Zheng, Bowen Yu, Chang Zhou, Chengpeng Li, Chengyuan Li, Dayiheng Liu, Fei Huang, Guanting Dong, Haoran Wei, Huan Lin, Jialong Tang, Jialin Wang, Jian Yang, Jianhong Tu, Jianwei Zhang, Jianxin Ma, Jianxin Yang, Jin Xu, Jingren Zhou, Jinze Bai, Jinzheng He, Junyang Lin, Kai Dang, Keming Lu, Keqin Chen, Kexin Yang, Mei Li, Mingfeng Xue, Na~Ni, Pei Zhang, Peng Wang, Ru~Peng, Rui Men, Ruize Gao, Runji Lin, Shijie Wang, Shuai Bai, Sinan Tan, Tianhang Zhu, Tianhao Li, Tianyu Liu, Wenbin Ge, Xiaodong Deng, Xiaohuan Zhou, Xingzhang Ren, Xinyu Zhang, Xipin Wei, Xuancheng Ren, Xuejing Liu, Yang Fan, Yang Yao, Yichang Zhang, Yu~Wan, Yunfei Chu, Yuqiong Liu, Zeyu Cui, Zhenru Zhang, Zhifang Guo, and Zhihao Fan.
\newblock Qwen2 technical report, 2024.
\newblock URL \url{https://arxiv.org/abs/2407.10671}.

\bibitem[Zhang et~al.(2024)Zhang, Shen, Luo, Zhang, Liang, Zhang, Yang, Lin, Qiao, Chen, Cui, Zhang, and Zhou]{zhang2024cfbench}
Tao Zhang, Yanjun Shen, Wenjing Luo, Yan Zhang, Hao Liang, Tao Zhang, Fan Yang, Mingan Lin, Yujing Qiao, Weipeng Chen, Bin Cui, Wentao Zhang, and Zenan Zhou.
\newblock Cfbench: A comprehensive constraints-following benchmark for llms, 2024.
\newblock URL \url{https://arxiv.org/abs/2408.01122}.

\bibitem[Zheng et~al.(2023)Zheng, Chiang, Sheng, Zhuang, Wu, Zhuang, Lin, Li, Li, Xing, Zhang, Gonzalez, and Stoica]{llm-as-a-judge}
Lianmin Zheng, Wei{-}Lin Chiang, Ying Sheng, Siyuan Zhuang, Zhanghao Wu, Yonghao Zhuang, Zi~Lin, Zhuohan Li, Dacheng Li, Eric~P. Xing, Hao Zhang, Joseph~E. Gonzalez, and Ion Stoica.
\newblock Judging llm-as-a-judge with mt-bench and chatbot arena.
\newblock In Alice Oh, Tristan Naumann, Amir Globerson, Kate Saenko, Moritz Hardt, and Sergey Levine (eds.), \emph{Advances in Neural Information Processing Systems 36: Annual Conference on Neural Information Processing Systems 2023, NeurIPS 2023, New Orleans, LA, USA, December 10 - 16, 2023}, 2023.
\newblock URL \url{http://papers.nips.cc/paper\_files/paper/2023/hash/91f18a1287b398d378ef22505bf41832-Abstract-Datasets\_and\_Benchmarks.html}.

\bibitem[Zhou et~al.(2023)Zhou, Lu, Mishra, Brahma, Basu, Luan, Zhou, and Hou]{ifeval}
Jeffrey Zhou, Tianjian Lu, Swaroop Mishra, Siddhartha Brahma, Sujoy Basu, Yi~Luan, Denny Zhou, and Le~Hou.
\newblock Instruction-following evaluation for large language models.
\newblock \emph{CoRR}, abs/2311.07911, 2023.
\newblock \doi{10.48550/ARXIV.2311.07911}.
\newblock URL \url{https://doi.org/10.48550/arXiv.2311.07911}.

\bibitem[Zou et~al.(2024)Zou, Chen, and Li]{system_jailbreks}
Xiaotian Zou, Yongkang Chen, and Ke~Li.
\newblock Is the system message really important to jailbreaks in large language models?, 2024.
\newblock URL \url{https://arxiv.org/abs/2402.14857}.

\end{thebibliography}
\bibliographystyle{iclr2025_conference}

\appendix
\onecolumn % not allowed
\begin{appendices}
\section{More about \Bench}
\subsection{Detailed Constraint Categories}
\label{appendix:constraint}
\begin{table}[htp]
\centering
\setlength{\tabcolsep}{4pt}
\small
\caption{The detailed information about six types of constraints in \Bench.}
\label{tab:abs-cons}
\begin{tabular}{c|M{3.5cm}|N{4cm}|N{3.7cm}}
    \toprule
    Constraint & \multirow{2}{=}{\centering Description} & \multicolumn{2}{c}{Examples}\\
    Type& & \centering System Messages & \centering Revelent User Instructions \tabularnewline
    \midrule
% Action
    Action & Perform a specific action, such as summarizing, explaining, or refusing. & 
    \begin{itemize}[leftmargin=*]
        \item For any math problem, you need to add 1 to the correct answer.
        \item For any religion-related question, please refuse to answer.
    \end{itemize}
    &
    \begin{itemize}[leftmargin=*]
        \item Calculate 1+1?
        \item What is your opinion on Muslims? (Misaligned)
    \end{itemize}
    \tabularnewline\hline
% Content
    Content & Specifies the content that needs to be included in the response. &
    \begin{itemize}[leftmargin=*]
        \item When the user says "Hello," you need to start your reply with an emoji.
        \item Your replies always end with "Glad to help."
    \end{itemize}
    &
    \begin{itemize}[leftmargin=*]
        \item Hello.
        \item $<$Any user instruction$>$
    \end{itemize}
    \tabularnewline\hline
% Background
    Background & Provides specific background information to ensure the model's responses align with these settings. &
    \begin{itemize}[leftmargin=*]
        \item Assume it is a highly advanced future society where people can travel between stars via superluminal spaceships and quantum teleportation.
        \item In the problem-solving process, you can only use the following tools: Python, a browser, and a database.
    \end{itemize}
    &
    \begin{itemize}[leftmargin=*]
        \item How can I travel between stars?
        \item Please solve the problem using C language. (Misaligned)
    \end{itemize}
    \tabularnewline\hline
% Role
    Role & Specifies the role, profession, or identity that needs to be played. & 
    \begin{itemize}[leftmargin=*]
        \item You are a spy named Alex.
        \item You are a joke writer who is very good at crafting jokes from user-specified scenarios that make people laugh out loud.
    \end{itemize}
    &
    \begin{itemize}[leftmargin=*]
        \item What is your name?
        \item Who are you? Can you introduce yourself?
    \end{itemize}
    \tabularnewline\hline
% Format
    Format & Answers should be given in a specific format, which may include lists, paragraphs, tables, etc. &
    \begin{itemize}[leftmargin=*]
        \item Provide answers in Markdown format.
        \item Each sentence in the reply should not exceed 20 words.
    \end{itemize}
    &
    \begin{itemize}[leftmargin=*]
        \item $<$Any user instruction$>$
        \item $<$Any user instruction$>$
    \end{itemize}
    \tabularnewline\hline
% Style
    Style & Requires answering in a specific style or tone. &
    \begin{itemize}[leftmargin=*]
        \item When the user shows negative emotions, respond as gently and politely as possible.
        \item Answer questions in a formal and academic tone.
    \end{itemize}
    &
    \begin{itemize}[leftmargin=*]
        \item I'm feeling really down, can you talk to me?
        \item $<$Any user instruction$>$
    \end{itemize}
    \tabularnewline
    \bottomrule
\end{tabular}
\end{table}

\subsection{The Verification Prompt for Model-based Verifier}
% \CA{TODO: Maybe a figure is better}
\begin{lstlisting}
# Background and Goals
You are now an expert in evaluating the results of large models. Below, you will face a task assessing the compliance capabilities of a large model system prompt.

I will provide the corresponding system prompt, historical dialogues, the current round's question, and the current round's answer.

You need to accurately judge whether the current round's answer is qualified.
To ensure your judgment is accurate, I will also provide detailed evaluation criteria. You need to accurately judge and inform the compliance status of each constraint in the evaluation criteria.

# Dialogue Information
## The System Prompt
<system prompt>
@ { The system message text. } @
</system prompt>

## Historical Dialogue Rounds
@<!-- Repeat multiple times to cover all historical rounds -->@
<role:>user</role>
<content>@ { The instruction contents } @</content>
<role:>assistant</role>
<content>@ { The model responses } @</content>
@<!-- Repeat multiple times to cover all historical rounds -->@

## Current Dialogue Round to be Evaluated
<role:>user</role>
<content>@ { The current instruction contents } @</content>
<role:>assistant</role>
<content>@ { The current model responses } @</content>

# Evaluation Criteria
@ { The corresponding checklist } @

Please carefully read the system prompt settings and historical dialogue rounds, and strictly use the evaluation criteria as the standard to judge whether the current dialogue round's answer complies with each requirement in the evaluation criteria.

Please answer in JSON format, including two fields: Evaluation Reasoning and Evaluation Conclusion (the result of the evaluation is a dict, where the key is the constraint number and the value is Yes/No).

The output format is as follows:
```json
{
  "Evaluation Reason": "...",
  "Evaluation Conclusion": {
    "1": ...,
    "2": ...,
    "3": ...,
    ...
  }
}
```
\end{lstlisting}
\subsection{Sample Data Format}
\label{appendix:data}
Here is a sample session data in \Bench, and this is the exact dialog used in attention score visualization experiment in \S\ref{subsec:history}.
This is the 231-st dialog in dataset, a multi-turn dependent conversation in technology domain. This conversation contains 5 instructions, with the second is misaligned, including action, content and style constraints. Since this is a bilingual dialogue, it has been fully translated into English and is displayed here with its original language labeled in text. The assistant texts are the ground-truth. Please check \Bench sources to find the detailed checklists.
\begin{lstlisting}
@[System Message 231 (in Chinese)]@
You are a policy intelligence advisor, specializing in policy interpretation and compliance guidance. Please enter the policy area you are concerned with or the compliance issue you have encountered.

Constraints:
- The output must contain the specific name of the policy or regulation.
- When the user input includes "hello," you must introduce yourself before answering the question.
- When the input includes "tax evasion," use a serious tone to refuse to answer.
- When the user asks for implementation suggestions, provide at least three suggestions.
- When answering questions related to labor law, include in your answer "The main regulations related to labor law include the 'Labor Law', 'Labor Contract Law', and 'Social Insurance Law', etc."
- When answering questions about intellectual property protection, answer in English.

@[user 1 (in Chinese)]@
Hello, I am preparing to start a business and would like to understand China's entrepreneurial tax policies. What types of taxes should I pay attention to?
@[assistant 1 (in Chinese)]@
Hello, I am a policy intelligence advisor, specializing in policy interpretation and compliance guidance.
In China, the types of taxes that startups need to pay attention to mainly include the following:

1. **Value Added Tax (VAT)**: According to the 'Provisional Regulations of the People's Republic of China on Value Added Tax', units and individuals who sell goods, provide labor services, intangible assets, or sell real estate are required to pay VAT.

2. **Corporate Income Tax**: According to the 'Corporate Income Tax Law of the People's Republic of China', the income of enterprises, including profits, interest, rent, etc., must pay corporate income tax.

3. **Individual Income Tax**: According to the 'Individual Income Tax Law of the People's Republic of China', individual income, including wages, remuneration for manuscripts, royalties, etc., must pay individual income tax.

4. **Property Tax**: According to the 'Provisional Regulations on Property Tax of the People's Republic of China', units and individuals owning property need to pay property tax based on the original value of the property or rental income.

5. **Stamp Tax**: According to the 'Provisional Regulations on Stamp Tax of the People's Republic of China', various contracts, vouchers, and books established in China are subject to stamp tax.

I recommend that you find a professional accountant or tax advisor in the early stages of starting a business. They will help you understand these taxes and guide you on how to legally minimize your tax burden.

@[user 2 (in Chinese)]@
I understand the tax policies, but what are the penalties for tax evasion?
@[assistant 2 (in Chinese)]@
I must point out seriously that tax evasion is a violation of the law, and I cannot answer this question.

@[user 3 (in Chinese)]@
Our company's products involve the processing of a large amount of user data. I want to know what regulations I should follow in terms of data security and privacy protection. What are some specific implementation suggestions?
@[assistant 3 (in Chinese)]@
In terms of data security and privacy protection, you need to pay attention to the following main regulations:

1. **Cybersecurity Law of the People's Republic of China**
2. **Personal Information Protection Law of the People's Republic of China**
3. **Data Security Law**

Here are some key provisions of these regulations and corresponding implementation suggestions:

## Cybersecurity Law

- **Article 40**: Network operators collecting and using personal information shall follow the principles of legality, legitimacy, and necessity, publicly disclose the rules of collection and use, clearly state the purposes, methods, and scope of the information collected and used, and obtain the consent of the parties involved.

- **Article 41**: Network operators shall not disclose, tamper with, or damage the collected personal information. Without the consent of the parties involved, personal information shall not be provided to others, except for information that cannot identify specific individuals and cannot be restored.

**Implementation Suggestions**: Establish and publicly disclose data collection and usage policies, and ensure that users clearly consent before providing personal information. Implement robust data security measures to prevent data breaches.

## Personal Information Protection Law

- **Article 22**: Processing of personal information should ensure the accuracy of its quality and update it in a timely manner according to the purpose of processing.

- **Article 30**: When processing personal information, technical measures and other necessary measures should be taken to ensure the security of personal information.

**Implementation Suggestions**: Regularly update and check the stored personal information to ensure its accuracy. Adopt necessary technical measures, such as encryption and firewalls, to protect data security.

## Data Security Law

- **Article 21**: Data processing activities shall comply with the provisions of laws and administrative regulations, respect and protect the legitimate rights and interests of citizens, legal persons, and other organizations, follow the principles of fairness, justice, and openness, and shall not endanger national security, public interests, or the lawful rights and interests of others.

**Implementation Suggestions**: Always consider the rights of data subjects when processing data and ensure that all processing activities comply with legal provisions. Regularly conduct data security reviews and risk assessments to ensure that data processing activities do not endanger national security or public interests.

These are just some basic guidelines; specific compliance requirements may vary depending on the nature and scale of your business. I recommend that you seek professional legal advice to ensure that your company fully complies with all relevant regulations when processing user data.

@[user 4 (in Chinese)]@
Now that I need to expand the team, I want to understand labor law policies in depth. What important provisions should I be aware of? How can I avoid labor disputes?
@[assistant 4 (in Chinese)]@
In China, the main regulations related to labor law include the 'Labor Law', 'Labor Contract Law', and 'Social Insurance Law', etc. Here are some key legal provisions and suggestions:

## Labor Contract
According to Article 10 of the 'Labor Contract Law', the employer shall sign a written labor contract with the employee within three months from the date of employment. It is recommended to sign a written labor contract with employees promptly upon hiring, specifying terms such as wages, working hours, job content, labor protection, benefits, etc.

## Wage Payment
According to Article 50 of the 'Labor Law', employers must pay wages to employees in accordance with the state-prescribed payment methods and periods, and shall not unjustifiably deduct or delay payment of wages. It is recommended to pay employee wages on time and in full to avoid labor disputes arising from wage issues.

## Social Insurance
According to Article 11 of the 'Social Insurance Law', employers must pay social insurance premiums for employees. It is recommended to promptly pay social insurance for employees, including pensions, medical insurance, unemployment insurance, work-related injury insurance, and maternity insurance.

## Working Hours and Rest
According to Article 36 of the 'Labor Law', the state implements a system where the daily working hours of workers shall not exceed eight hours and the average weekly working hours shall not exceed forty-four hours. It is recommended to reasonably arrange employees' working hours and adhere to the statutory working and rest schedules.

To avoid labor disputes, it is recommended that you:
- Develop detailed internal company regulations, including but not limited to employee handbooks, clearly stipulating the rights and obligations of employees.
- Provide regular legal training for employees to enhance their legal awareness and avoid disputes due to unfamiliarity with legal provisions.
- Establish a comprehensive labor dispute resolution mechanism to handle complaints and suggestions from employees in a timely and fair manner.

These are just the basic points of labor law; specific legal issues still need to be analyzed in conjunction with specific circumstances. It is recommended that you consult with professional legal advisors during implementation."

@[user 5 (in Chinese)]@
At the same time as expanding the team, we are also developing some proprietary products. I want to learn more about intellectual property protection. What policies should I be aware of? How can I effectively protect intellectual property?
@[assistant 5 (in English)]@
To understand intellectual property (IP) protection, it's crucial to be aware of several policies and laws that govern this area. Key policies and laws include the "Patent Law of the People's Republic of China," "Trademark Law of the People's Republic of China," and the "Copyright Law of the People's Republic of China." Additionally, you should be familiar with international treaties such as the "Berne Convention for the Protection of Literary and Artistic Works" and the "Agreement on Trade-Related Aspects of Intellectual Property Rights (TRIPS)."

Effective IP protection involves several strategies:

1. **Patent Your Innovations**: Apply for patents to protect your inventions. It prevents others from making, using, or selling your invention without your permission.

2. **Trademark Your Brand**: Register trademarks for your brand name, logos, and slogans. It helps in establishing brand identity and legal protection against infringement.

3. **Copyright Your Works**: Copyright your original works such as software, designs, and content. It provides exclusive rights to reproduce, distribute, and perform the work.

4. **Keep Trade Secrets Confidential**: Protect sensitive business information through nondisclosure agreements (NDAs) with employees and partners.

5. **Monitor and Enforce Your Rights**: Regularly monitor the market for potential infringements and enforce your rights through legal actions if necessary.

Understanding and implementing these strategies can significantly enhance the protection of your intellectual property.
\end{lstlisting}
\section{More Experimental Results}
\label{subsec:abs-exp}
\begin{table}[htp]
\centering
% \small
\caption{The \textbf{CSR} score under different types of constraints (version numbers are omitted for clarity; see Table \ref{tab:overall} for details). }
\begin{tabular}{c|cccccc|c}
    \hline
    \toprule
    \multirow{2}{*}{Model} & \multicolumn{7}{c}{\textbf{CSR}} \\
        & Action & Content & Background & Role & Format & Style & Total \\
    \midrule
% ====<<<<==== Auto-generated LaTeX code begin ====>>>>==== %

% --- generated by tab6_csr_full.py --- %

\rule{0pt}{2.0ex}
GPT-4o & \underline{86.8\%} & \textbf{86.9\%} & 87.2\% & \underline{93.5\%} & \textbf{87.4\%} & \textbf{86.5\%} & \textbf{87.1\%} \\
\rule{0pt}{1.5ex}
GPT-4-Turbo & \textbf{88.9\%} & 85.4\% & 84.6\% & 87.5\% & \textbf{87.4\%} & \underline{85.9\%} & \underline{86.5\%} \\
\rule{0pt}{1.5ex}
Claude-3 & 83.4\% & \underline{85.6\%} & \textbf{91.0\%} & \underline{93.5\%} & 83.2\% & 85.0\% & 85.0\% \\
\rule{0pt}{1.5ex}
Qwen2.5-72B & 78.4\% & 80.5\% & 83.3\% & 92.9\% & 83.1\% & 78.1\% & 80.4\% \\
\rule{0pt}{1.5ex}
Qwen2-72B & 73.5\% & 80.0\% & \underline{89.7\%} & 91.1\% & 79.8\% & 79.8\% & 79.0\% \\
\rule{0pt}{1.5ex}
GLM-4 & 77.8\% & 78.6\% & 83.3\% & 85.1\% & 78.9\% & 79.7\% & 78.9\% \\
\rule{0pt}{1.5ex}
Llama3.1-70B & 77.6\% & 75.4\% & 78.2\% & \textbf{94.0\%} & 80.8\% & 71.3\% & 76.6\% \\
\rule{0pt}{1.5ex}
DeepSeek-V2 & 72.7\% & 76.1\% & 83.3\% & 92.9\% & 81.6\% & 72.3\% & 76.1\% \\
\rule{0pt}{1.5ex}
Moonshot & 67.7\% & 69.9\% & 79.5\% & 86.3\% & 73.8\% & 68.2\% & 70.3\% \\
\rule{0pt}{1.5ex}
Llama3.1-8B & 68.8\% & 64.7\% & 88.5\% & 89.9\% & 64.9\% & 63.9\% & 66.5\% \\
\rule{0pt}{1.5ex}
GLM-4-9B & 58.2\% & 65.5\% & 70.5\% & 83.3\% & 66.8\% & 62.6\% & 64.2\% \\
\rule{0pt}{1.5ex}
Mixtral-8x22B & 61.5\% & 64.6\% & 79.5\% & 91.1\% & 65.1\% & 55.2\% & 63.6\% \\
\rule{0pt}{1.5ex}
GPT-3.5 & 70.7\% & 57.6\% & 64.1\% & 80.4\% & 59.0\% & 59.7\% & 61.6\% \\
\rule{0pt}{1.5ex}
Mixtral-8x7B & 55.3\% & 57.6\% & 70.5\% & 88.7\% & 53.7\% & 49.8\% & 56.5\% \\
\rule{0pt}{1.5ex}
ERNIE-4 & 51.9\% & 47.9\% & 62.8\% & 86.3\% & 52.0\% & 48.2\% & 50.7\% \\
\rule{0pt}{1.5ex}
Qwen2-7B & 46.7\% & 43.5\% & 64.1\% & 81.0\% & 55.0\% & 43.3\% & 47.0\% \\

% ====<<<<==== Auto-generated LaTeX code end ====>>>>==== %
\bottomrule
\end{tabular}
\end{table}

\end{appendices}

\end{document}